%% file: main.tex
\documentclass[10pt,twocolumn,letterpaper]{article}

\usepackage[pagenumbers]{cvpr} 


\definecolor{cvprblue}{rgb}{0.21,0.49,0.74}
\usepackage[pagebackref,breaklinks,colorlinks,allcolors=cvprblue]{hyperref}
\usepackage{multirow,makecell,stfloats}
\usepackage{lipsum}
\usepackage{hyperref}
\usepackage{marvosym}

\def\paperID{14676} 
\def\confName{CVPR}
\def\confYear{2025}

\def\ourmethod{HoloDrive}
\title{\ourmethod: Holistic 2D-3D Multi-Modal Street Scene Generation for Autonomous Driving}

\author{
\begin{tabular}{cccccccc}
Zehuan Wu$^{1*}$$^{\href{mailto:wuzehuan@sensetime.com}{\textrm{\Letter}}}$ & 
Jingcheng Ni$^{1*}$ & 
Xiaodong Wang$^{3*}$ & 
Yuxin Guo$^{1*}$ & 
Rui Chen$^{1*}$ & \\
& Lewei Lu$^{1}$ &
Jifeng Dai$^{2,4}$  & 
Yuwen Xiong$^{2}$$^{\href{mailto:yuwen.xiong.cs@gmail.com}{\textrm{\Letter}}}$
\end{tabular}
\\
$^1$Sensetime Research
$^2$Shanghai Artificial Intelligence Laboratory \\
$^3$Peking University
$^4$Tsinghua University\\
}

\begin{document}

\input{sec/0_abstract}    
\newcommand\blfootnote[1]{%
\begingroup
\renewcommand\thefootnote{}\footnote{#1}%
\addtocounter{footnote}{-1}%
\endgroup
}
\blfootnote{$^{*}$equal contribution}
\input{sec/1_intro}
\input{sec/2_related}
\input{sec/3_method}
\input{sec/4_exp}
\input{sec/5_conclusion}

{
    \small
    \bibliographystyle{ieeenat_fullname}
    \bibliography{main}
}
\input{sec/X_suppl}

\end{document}

%% file: sec/0_abstract.tex

\twocolumn[{%
\renewcommand\twocolumn[1][]{#1}%
\maketitle
\vspace{-5mm}
\includegraphics[width=\textwidth]{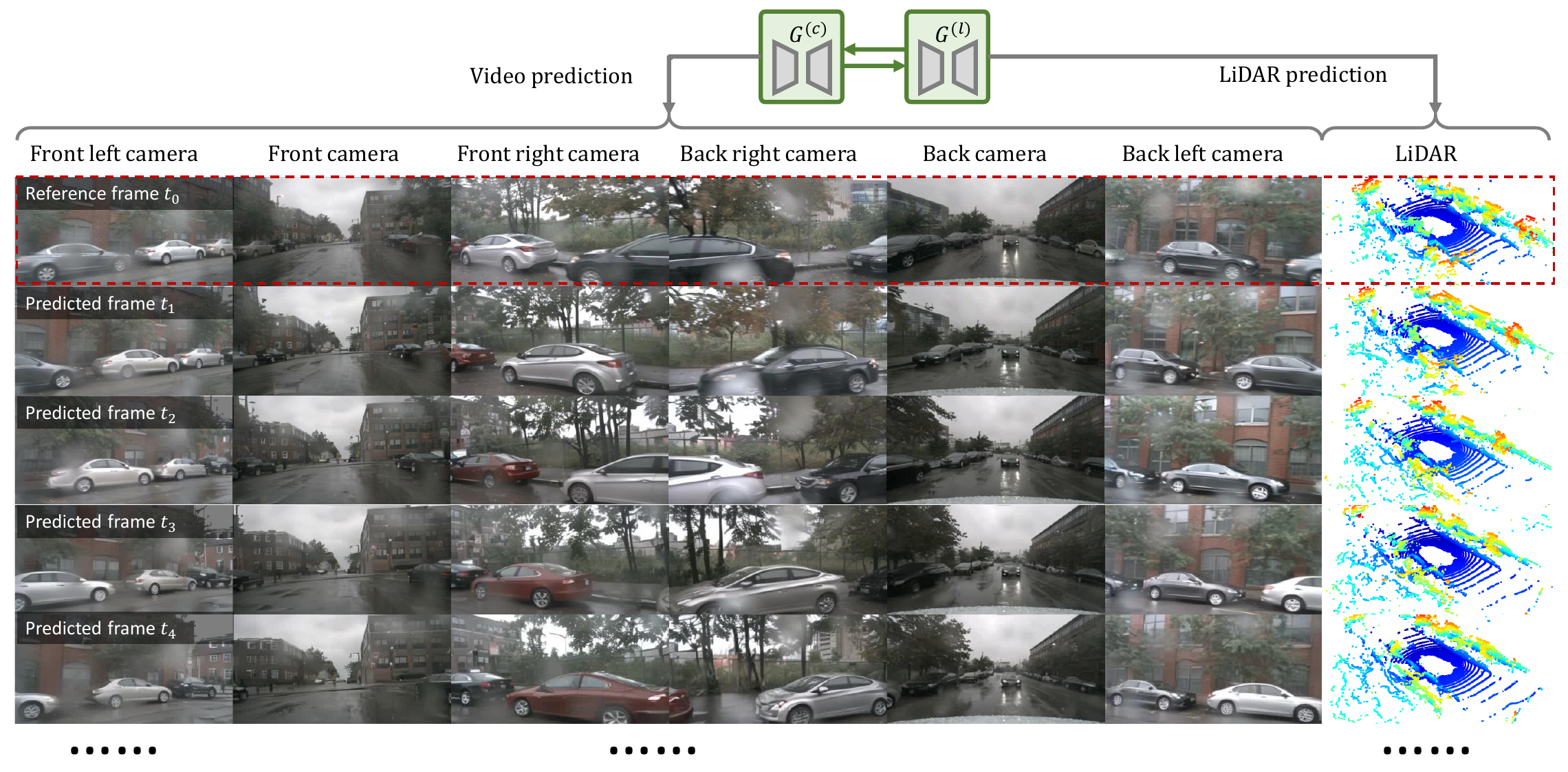}
\vspace{-7mm}
\captionof{figure}{Our pipeline \ourmethod~can jointly generate realistic street scene video of surround-view cameras and LiDAR point cloud.}
\vspace{6mm}
\label{fig:teaser}
}]

\begin{abstract}
Generative models have significantly improved the generation and prediction quality on either camera images or LiDAR point clouds for autonomous driving.
However, a real-world autonomous driving system uses multiple kinds of input modality, usually cameras and LiDARs, where they contain complementary information for generation, while existing generation methods ignore this crucial feature, resulting in the generated results only covering separate 2D or 3D information.
In order to fill the gap in 2D-3D multi-modal joint generation for autonomous driving, in this paper, we propose our framework, \emph{HoloDrive}, to jointly generate the camera images and LiDAR point clouds.
We employ BEV-to-Camera and Camera-to-BEV transform modules between heterogeneous generative models, and introduce a depth prediction branch in the 2D generative model to disambiguate the un-projecting from image space to BEV space, then extend the method to predict the future by adding temporal structure and carefully designed progressive training.
Further, we conduct experiments on single frame generation and world model benchmarks, and demonstrate our method leads to significant performance gains over SOTA methods in terms of generation metrics.
\end{abstract}

%% file: sec/1_intro.tex
\section{Introduction}

\label{sec:intro}
Generative models have gained significant attention for their ability to understand data distributions and create content, making notable strides in areas such as image~\citep{peebles2023scalable, saharia2022photorealistic, chang2022maskgit} and video generation~\citep{blattmann2023videoldm}, 3D object generation~\citep{xiong2023ultralidar, khurana20234docc}, and editing~\citep{huang2024subjectdrive}.
In the context of simulation, generative models have shown remarkable potential for creating realistic scenarios, which are crucial for training and evaluating safety-critical embodied agents like autonomous vehicles~\citep{hu2023gaia, wang2024driving}.
This capability reduces the need for expensive manual modeling of the real world, facilitating extensive closed-loop training and scenario testing.
Furthermore, world models are gradually being explored to understand and predict the dynamic nature of the real world, which is crucial for simulation scenes and video generation.

Despite the advancements in conditional image and video generation for autonomous driving, existing approaches primarily focus on a single modality, utilizing either 2D data~\citep{wang2024driving,zhao2024drivedreamer} or 3D data~\citep{xiong2023ultralidar,zhang2023copilot4d}.
A truly capable autonomous driving system, however, usually integrates multiple sensors, including both cameras and LiDARs.
Cameras provide rich texture and semantic information, while LiDARs offer precise 3D geometric details.
The combination of these two modalities enhances perception accuracy, as they are complementary~\citep{liang2022bevfusion,li2022deepfusion}.
The exploration of joint modality generation is still very preliminary at present.
BEVWorld~\citep{zhang2024bevworld} has made some explorations, but the quality and controllability of generation are still difficult to compare with the SOTA methods in single modality.

We propose a holistic 2D-3D generation framework for autonomous driving, \textbf{\ourmethod}, which unifies 2D and 3D generation for street-view autonomous driving data into a single and effective framework.
\ourmethod~can jointly generate both multi-view camera and LiDAR data, as illustrated in Figure~\ref{fig:teaser}.
Our framework extends state-of-the-art 2D and 3D generation models, enabling the generation of realistic street scenes with text and bounding box/map conditions.

To achieve joint 2D and 3D generation, we first introduce a depth prediction branch in the 2D generative model, with supervision naturally derived from the 3D LiDAR.
We further employ an efficient BEV-to-Camera transformation based on this depth prediction to align the 3D and 2D spaces, and also a Camera-to-BEV module that introduces rich 2D semantic priors to 3D space.
These cross-modal structures facilitate effective information exchange between the two modalities during the generation process and making the entire model end-to-end trainable.
We apply the joint pipeline on both single frame and video generation tasks with progressive training, combined with extra multi-task learning on the video domain for a smooth transition across training stages.

We conduct experiments on the NuScenes dataset~\citep{caesar2020nuscenes}, which provides information including paired multi-view camera images, LiDAR point clouds, text descriptions, and map layouts.
Our results show that integrating joint 2D-3D modeling, \ourmethod~achieves state-of-the-art performance in generating both single-frame and sequential data of multi-view camera images and LiDAR point clouds.
The main contributions of this paper can be summarized as follows:

\begin{itemize}[leftmargin=1cm]
  \item We present a novel framework, \ourmethod, to jointly generate multi-view camera images and LiDAR point clouds that are consistent in 2D and 3D space, given text and layout conditions.
  \item We propose to add extra depth supervision to the 2D generation and apply an efficient Camera-to-BEV transformation model to align the 2D and 3D spaces, enhancing joint 2D-3D generative modeling, and further extending the joint modeling to video generation.
  \item Our method shows superior generation quality, faithfully following given conditions, as well as 2D-3D consistency, achieving state-of-the-art performance for both single-frame and video generation.
\end{itemize}

%% file: sec/2_related.tex
\section{Related Work}

\subsection{Image Generation}
Image generation is one of the most basic topics in generative modeling and various methods have been explored.
Among them, diffusion models, which model the image generation via a reverse iterative stochastic process, raise more and more attention to competitive training stability and generation quality.
The reasons behind are carefully design choices in diffusion models, including reducing the prediction resolution by auto-encoder~\citep{rombach2021highresolution} or cascade model~\citep{saharia2022photorealistic}, better noise scheduler~\citep{song2020score, karras2022elucidating, ma2024sit}, classifier-free guidance~\citep{ho2021classifier} for control capability and so on.
More recently, several works~\citep{peebles2023scalable, li2024hunyuan, lu2024fit} manage to transfer the scaling ability of the Transformer~\citep{vaswani2017attention} to diffusion models that has shown priority in the NLP area.

Compared with natural images, there exists inherent differences, i.e., regular scene structures and diverse objects in the autonomous driving (AD) area.
To compensate for the differences, layout information is utilized to guide generation.
For instance, BEVGen~\citep{swerdlow2024street} refers to 3D information by projecting all layouts into BEV space.
Conversely, BEVControl~\citep{yang2023bevcontrol} begins from projecting 3D coordinates to image views to construct 2D geometric guidance, and MagicDrive~\citep{gao2023magicdrive} combines the advantage of both methods.
Recently, Drive-WM~\citep{wang2024driving} transfers the pixel-level layout information to latent space and resorts to a unified embedding to attend to them.
Our method makes further improvements by introducing point-cloud synergy.

\subsection{LiDAR Generation}
LiDAR point cloud generation has been explored in recent years, a task belonging to 3D point cloud generation.
Early works utilized the variational autoencoder (VAE)~\citep{kingma2013auto} or generative adversarial network (GAN)~\citep{goodfellow2014generative} on point cloud to enable unconditional generation~\citep{caccia2019lidarvae,sallab2019lidar}.
LiDARGen~\citep{zyrianov2022lidargen} leverages a score-matching energy-based model and denoise from pure noise into point clouds on the equirectangular view.
To better maintain the structure and semantic information of LiDAR scenes, UltraLiDAR~\citep{xiong2023ultralidar} first proposes to utilize a discrete representation to model the distribution of LiDAR.
They train a LiDAR VQ-VAE~\citep{van2017vqvae} to learn discrete representations and then leverage a bidirectional transformer~\citep{chang2022maskgit} to learn the joint distribution of discrete tokens of LiDAR scenes.
Regarding point cloud forecasting, some methods exploit past LiDAR scans to predict future point clouds, modeling the temporal dynamics based on LSTM~\citep{weng2021SPFNet}, stochastic sequential latent models~\citep{weng2022s2net}, or 3D spatiotemporal convolutional networks~\citep{mersch2022ST3DCNN}.
4D-Occ~\citep{khurana20234docc} chooses to forecast a generic future 3D occupancy-like quantity, instead of directly predicting future point clouds.
Copilot4D~\citep{zhang2023copilot4d} explores discrete diffusion models in future LiDAR prediction and combined training objectives including individual frame prediction, future prediction, and joint modeling.
RangeLDM~\citep{hu2024rangeldm} learns to generate by denoising the latent of LiDAR range images, and those images are projected from point clouds via Hough Voting to ensure high quality representation.
However, these methods only consider the priors of LiDAR point clouds, lacking semantic and perceptual information.
In this work, our proposed \ourmethod~utilizes information from both 2D images and 3D point cloud priors, facilitating the generation of high-quality point clouds.

\subsection{Joint Generation}
BEVWorld~\citep{zhang2024bevworld} first works on camera and LiDAR joint generation, and proposes a unified BEV latent representation of both camera and LiDAR leveraging ray cast module inside the latent autoencoder, then generates by denoising the unified BEV latent.
However, this newly designed latent space has not been trained with large-scale data, so the image generation quality is still hard to match with those methods that are fine-tuned on large-scale pre-trained models like SD.
Our proposed \ourmethod~building on the effective use of the ability from pre-trained image generation models, achieves 2D-3D joint generation and reaches the state-of-the-art (SOTA) level in terms of generation quality.

\subsection{Predictive World Model}
Predictive World Model, leveraging a generalized predictive model to learn from sequential data, serves as one of the potential ways to reproduce the great success of LLMs~\citep{touvron2023llama} in the vision area.
In the vision domain, predictive models can be regarded as a special form of video generation, with past observations as guidance.
Further narrowing down to the field of AD, DriveGAN~\citep{kim2021drivegan} and GAIA-1~\citep{hu2023gaia} learn a generalized driving video predictor with action-conditioned video diffusion models.
DriveDreamer~\citep{wang2023drivedreamer} introduces extra 3D conditions and a progressive training strategy.
GenAD~\citep{yang2024genad} enlarges the model by building a larger dataset.
To further improve the prediction ability, ADriver-I~\citep{jia2023adriver} utilizes LLM-generated abstract signals, e.g., action and speed.
While the aforementioned methods learn from monocular videos, most recently, Drive-WM~\citep{wang2024driving} and DriveDreamer-2~\citep{zhao2024drivedreamer} extend the learning resource to multi-view videos.
Despite the competitive results achieved by these methods, it remains unknown whether these models are aware of the 3D world. In this work, we pioneer a path towards cooperative generation of multi-view videos and point clouds.

%% file: sec/3_method.tex
\section{Method}

\begin{figure*}[t]
  \centering
  \includegraphics[width=\textwidth]{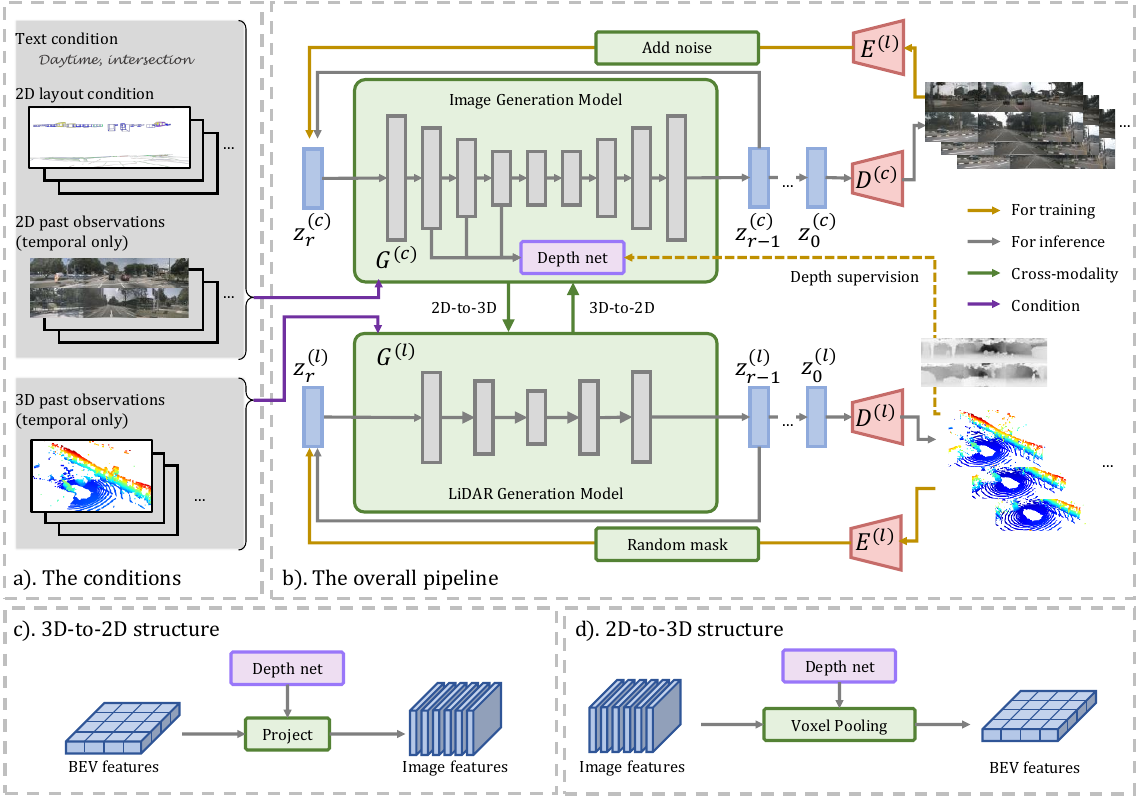}
  \captionsetup{justification=raggedright,singlelinecheck=false}
  \caption{\textbf{Overview of the proposed pipeline.} a). The conditions used by our pipeline. b).The overall joint training and inference pipeline. c). The structure to convert BEV features for the image generation model. d). The structure to convert image features for the LiDAR generation model.}
  \label{fig:pipeline}
\end{figure*}

{Fig.~\ref{fig:pipeline} illustrates the overview of the proposed pipeline, which jointly predicts the multi-view video and future LiDAR points.
In addition to basic 2D and 3D generation models, two novel cross-modal structures, 2D-to-3D and 3D-to-2D structures, are proposed to achieve interactions between the two modalities and jointly improve the quality of video (or image) and LiDAR generation.
For multi-modal data and models, superscript $^{(c)}$ indicates camera, and superscript $^{(l)}$ indicates LiDAR.

\subsection{Multi-view Image Generation}

The basic image generation pipeline in our method follows SD 2.1~\citep{rombach2021highresolution}.
Given the original image $o^{(c,k)}\in \mathbb{R}^{H^c\times W^c\times 3}$, $k$ for the view index, $H^c$, $W^c$ for the image height and width, we get the image latent $z^{(c,k)}=E^{(c)}(o^{(c,k)})$, where $E^{(c)}$ is the VAE encoder.
It iteratively denoises from a random Gaussian noise $z^{(c)}_R$ for $R$ steps with a U-Net model $G^{(c)}$ into a clean image latent $z^{(c)}_0$.

\noindent\textbf{Cross-view attention.}
Following Drive-WM~\citep{wang2024driving}, cross-view attention blocks are inserted after each spatial attention block in the diffusion U-Net for multi-view consistency.
The cross-view attention block takes the output of the U-Net spatial blocks, and applies self-attention across different views, then merges the output into its input by a learnable mixer.

\noindent\textbf{Conditions.}
We use simple scene descriptions as text condition $P$ and affect the model through cross-attention.
The projected 3D box $B^{(c)}\in \mathbb{R}^{H^c\times W^c\times 3}$ and projected HD map condition $H^{(c)}\in \mathbb{R}^{H^c\times W^c\times 3}$ are concatenated on the channel dimension as $e^{(c)}=[B^{(c)},H^{(c)}]$, and then injected into the model following T2I-Adapter~\citep{mou2023t2i} for flexibility.

Denoting $z^{(c)}_r(\epsilon)=\sqrt{\bar{\alpha}_r}z^{(c)}_0 + \sqrt{1-\bar{\alpha}_r}\epsilon$ as noisy latents, where $r$ is a timestep, $\epsilon\sim{\mathcal{N}(0, I)}$ is Gaussian noise, $\bar{\alpha}_r$ is hyper-parameter, we train the model with the training objective
\begin{equation}
    \mathcal{L}^{(c)} =
    \mathbb{E}_{{z^{(c)}_0},\epsilon\sim{\mathcal{N}(0, I)},r}\left[ \Vert
    \epsilon - G^{(c)}_\theta(z^{(c)}_r(\epsilon),r,P,e^{(c)})
    \Vert^2_2 \right].
\label{eq:img_diff}
\end{equation}

\subsection{LiDAR Generation}
Our method learns to generate LiDAR point clouds using discrete representations~\citep{van2017vqvae}.
We train a VQ-VAE-like tokenizer following UltraLiDAR~\citep{xiong2023ultralidar}.
Given a LiDAR points cloud observation $o^{(l)}$, we utilize an encoder-decoder model to quantify and reconstruct it.
The encoder $E^{(l)}$ is a PointNet~\citep{qi2017pointnet} followed by several Swin Transformer blocks~\citep{liu2021swin}, which converts point clouds into BEV latent features, and the output of encoder $z^{(l)}=E^{(l)}(o^{(l)})$ goes through a quantization layer to obtain discrete tokens $\hat{z}^{(l)}$.
The decoder $D^{(l)}$ has several Swin Transformer blocks and additional differentiable depth rendering branch~\citep{zhang2023copilot4d} for voxel reconstruction.
During inference, when discrete tokens are decoding into point clouds, spatial skipping~\citep{zhang2023copilot4d} is used to speed up sampling.

We then train a generative model that can generate diverse LiDAR point clouds.
Different from UltraLiDAR~\citep{xiong2023ultralidar} that only generates LiDAR point clouds unconditionally, we propose a generative model conditioned on multi-channel BEV features  $e^{(l)}$.
The BEV condition features can either be the 3D box and HD map conditions directly projected from the dataset annotation, or the cross-modal conditions converted from the feature map in the 2D generation network.
We adopt a training paradigm similar to MaskGIT~\citep{chang2022maskgit} but employ a U-Net like transformer $G^{(l)}$ to perceive multi-scale features.
Given the ground-truth LiDAR point cloud $o^{(l)}$, we tokenize it into a sequence of BEV tokens $Z^{(l)}=(z_1^{(l)}, z_2^{(l)}, \cdots, z_N^{(l)})$, where $N$ denotes the total number of tokens. 
During training, we mask a portion of tokens with mask ratio $\gamma(u)$ by replacing them with a special $[\texttt{MASK}]$ token according to a cosine mask scheduler $\gamma$ and $u\sim \texttt{Uniform}(0,1)$.
The training objective is defined to reconstruct the original inputs with a cross-entropy loss
\begin{equation}
    \mathcal{L}^{(l)} = 
    -\mathop{\mathbb{E}}
    _{Z\sim \mathcal{D}} 
    \left[ \sum_{{\forall}i \in [1, N], m_i=1} \log P\left(y_i\left|Z^{(l)}_{\overline{M}}, e^{(l)}\right.\right)  \right],
\label{eq:lidar}
\end{equation}
in which $Z^{(l)}_{\overline{M}}$ is the BEV tokens masked by $\overline{M}$ and  $P\left(y_i\left|Z^{(l)}_{\overline{M}}, e^{(l)}\right.\right)$ denotes the output probabilities of the transformer.
The transformer $G^{(l)}$ has two directions to model the distribution of LiDAR tokens and consists of Swin Transformer blocks~\citep{liu2021swin}.
We adopt a LiDAR token sampling algorithm similar to the sampling process in MaskGIT~\citep{chang2022maskgit}, in which the number of masked tokens $n = \lceil \gamma(t/T)N\rceil$ at iteration $t$ follows a mask scheduler $\gamma$, and $T$ is the total number of iterations.
Eventually, the generated tokens $\hat{Z}^{(l)}$ are decoded into LiDAR point clouds through the tokenizer decoder $D^{(l)}$ with depth rendering.

\subsection{Joint Generation of Camera and LiDAR}

\begin{figure}[t]
  \centering
  \includegraphics[width=\linewidth]{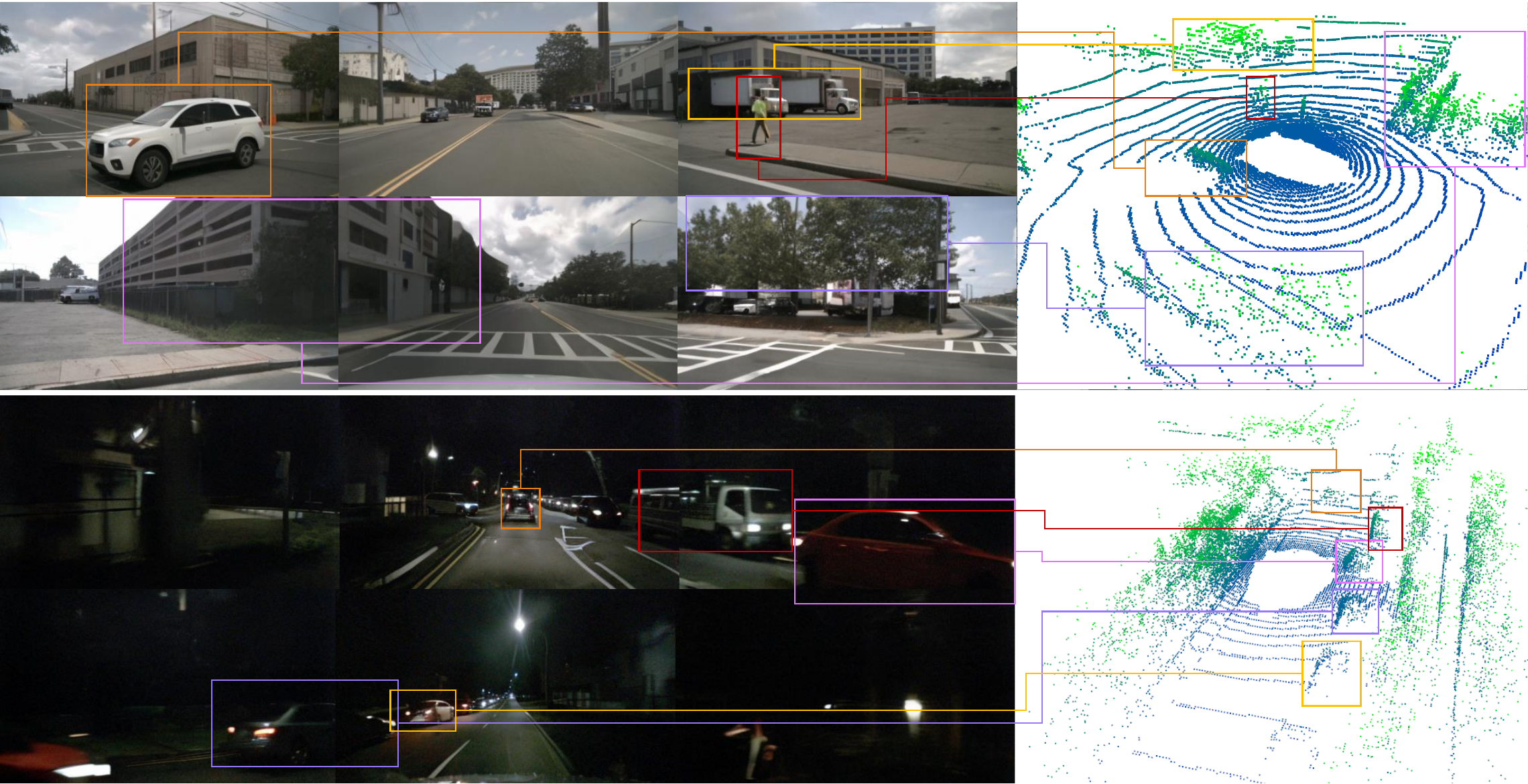}
  \captionsetup{justification=raggedright,singlelinecheck=false}
  \caption{One visual result of our joint 2D-3D generation. As indicated by the colored boxes and lines, our generation results exhibit high consistency across the two modalities.}
  \label{fig:visualization}
\end{figure}

\begin{table}[bp]
  \centering
  \begin{tabular}{c|ccc}
  \toprule
  \textbf{Method} & \textbf{FID$\downarrow$} & \textbf{\makecell[c]{BEVFusion\\mAP$_\text{obj}$$\uparrow$}} & \textbf{\makecell[c]{BEVFormer\\mAP$_\text{obj}$$\uparrow$}} \\
  \midrule
  BEVGen \cite{swerdlow2024street} & 25.54 & - & - \\
  GliGEN \cite{li2023gligen} & - & - & 15.42 \\
  BEVControl \cite{yang2023bevcontrol} & 24.85 & - & 19.64 \\
  BEVWorld \cite{zhang2024bevworld} & 19.0 & - & - \\
  MagicDrive \cite{gao2023magicdrive} & 16.20 & 12.30 & - \\
  Drive-WM \cite{wang2024driving} & 12.99 & - & \textbf{20.66} \\
  \midrule
  \ourmethod~(ours) & \textbf{10.64} & \textbf{14.06} & 19.98 \\
  \bottomrule
  \end{tabular}
  \caption{Image generation comparison.}
  \label{tab:image-generation-results}
\end{table}

As illustrated in Fig.~\ref{fig:pipeline} (c) and (d), two structures are employed for interactions between the 2D and 3D models: two unidirectional cross-modal transformation modules and a depth supervision module.
The former aims at improving the quality of generated elements and cross-modal consistency, while the later enables better 3D perception.

\noindent\textbf{Depth supervision.}
We follow BEVDepth~\citep{li2022bevdepth} to estimate depth using image features extracted from U-Net down blocks.
All the output features of the down blocks are resized to $\frac{H^c}{L}\times \frac{W^c}{L}$ then be concatenated, $L$ for the scale level of the VAE, usually be 8.
The output of this net is $F^{(c)}_d\in \mathbb{R}^{{\frac{H^c}{L}\times \frac{W^c}{L}\times D}}$, $D$ for the count of depth bins.
Given depth prediction and projected point clouds as ground truth, we calculate depth loss $\mathcal{L}^{(d)}$, which is simply a Cross Entropy loss.

\noindent\textbf{3D → 2D.}
Our 3D-to-2D module projects 3D features onto the 2D perspective view. Specifically, we first create a frustum-shaped point cloud $p_{(c)}\in \mathbb{R}^{\frac{H^c}{L}\times \frac{W^c}{L}\times D\times 3}$ for each camera.
Each point is calculated from its image space homogeneous coordinate times the actual distance of the depth bin.
By solving the equation
\begin{align}
    p^T_{(c)}=K_{(c)}\cdot (R_{(l)\to (c)}\cdot p^T_{(l)}+T_{(l)\to (c)}),
\end{align}
where $K_{(c)}$ refers to the matrix of camera intrinsic parameters, $R_{(l)\to (c)}$  the rotation matrix from the LiDAR space to the camera space, $T_{(l)\to (c)}$ the translation vector from the LiDAR space to the camera space, and $p_{(l)}$ the frustum-shaped point cloud in the LiDAR space.
Then we sample the hidden states of the down-blocks of the LiDAR generation model with $p_{(l)}$ and calculate the summation along the depth dimension weighted by the $F^{(c)}_d$, and finally arrive at $e^{(\texttt{proj})}$.
A lightweight adapter~\citep{mou2023t2i} is employed to inject the sampled features.
Similarly to the 2D-to-3D side, we concatenate the projected features $e^{(\texttt{proj})}$ and the 2D condition features $[B^{(c)}, H^{(c)}]$ into an unified 2D condition features $e^{(c)}$=$[B^{(c)}, H^{(c)}, e^{(\texttt{proj})}]$ as an updated version of $e^{(c)}$ in Eq.~\ref{eq:img_diff}. 

\noindent\textbf{2D → 3D.}
We propose a novel 2D-to-3D module that aggregates 2D prior knowledge from the 2D multi-view generative model into 3D space, which provides semantic information of the surrounding environment. We use voxel pooling following BEVDepth~\citep{li2022bevdepth} to convert the multi-view intermediate features from the 2D model, i.e., noisy latent features.
During training, following Eq.~\ref{eq:img_diff}, we obtain the multi-view intermediate features $F_r^{(c)}$ from U-Net blocks for timestep $r$ given 2D conditions.
Using the $F^{(c)}_d$ as weights, the features in the image space are converted to the BEV space as embedding $e^{(l)}$ through voxel pooling.

\begin{table}[bp]
  \centering
  \begin{tabular}{c|cccc}
  \toprule
  \textbf{Method} & \textbf{\makecell[c]{Multi-\\view}} & \textbf{\makecell[c]{Pre-\\train}} & \textbf{FID$\downarrow$} & \textbf{FVD$\downarrow$} \\
  \midrule
  DriveGAN \cite{kim2021drivegan} & & - & 73.4 & 502.3 \\
  DriveDreamer \cite{wang2023drivedreamer} & & SD & 52.6 & 452.0 \\
  BEVWorld \cite{zhang2024bevworld} & $\surd$ & - & 37.4 & 154.0 \\
  Drive-WM \cite{wang2024driving} & $\surd$ & SD & 15.8 & 122.7 \\
  DriveDreamer-2 \cite{zhao2024drivedreamer} & $\surd$ & SVD & 18.4 & \textbf{74.9} \\
  \midrule
  \ourmethod~(ours) & $\surd$ & SD & \textbf{13.6} & 103.0 \\
  \bottomrule
  \end{tabular}
  \caption{Video generation comparison.}
  \label{tab:video-main}
\end{table}

\noindent\textbf{Joint training \& inference.}
We optimize the joint training stage based on the summation of all the training objectives with balancing weights $\lambda_{l}$, $\lambda_{c}$ and  $\lambda_{d}$:
\begin{equation}
    \mathcal{L} = \lambda_{l}\mathcal{L}^{(l)} +  \lambda_{c}\mathcal{L}^{(c)} +  \lambda_{d}\mathcal{L}^{(d)}.
\end{equation}

\begin{table*}[t]
  \centering
  \begin{tabular}{c|ccccc}
  \toprule
  \textbf{Method} & \textbf{Chamfer$\downarrow$} & \textbf{L1 Mean$\downarrow$} & \textbf{\makecell[c]{AbsRel L1\\Mean(\%)$\downarrow$}} & \textbf{L1 Med$\downarrow$} & \textbf{\makecell[c]{AbsRel L1\\Med(\%)$\downarrow$}} \\
  \midrule
  UltraLiDAR \cite{xiong2023ultralidar} (Uncond) & 14.54 & 4.52 & 43.40 & 1.14 & 13.21 \\
  UltraLiDAR \cite{xiong2023ultralidar} (Cond) & 8.45 & 3.06 & 27.31 & 0.69 & 8.07 \\
  \ourmethod~(ours) & \textbf{7.61} & \textbf{2.37} & \textbf{16.89} & \textbf{0.46} & \textbf{5.71} \\
  \bottomrule
  \end{tabular}
  \caption{Single-frame LiDAR generation comparison.}
  \label{tab:lidar-gen-comp}
\end{table*}

\subsection{Temporal Modeling}

\noindent\textbf{Temporal generator architecture.}
To build a world model with multi-modal video generation, We model temporal information by following the method of Drive-WM~\citep{wang2024driving} that inserts temporal attention layers after the spatial attention layers. We also follow the design of Copilot4D~\citep{zhang2023copilot4d} to introduce a causal mask on the 3D video generator. 

\noindent\textbf{Joint world model.} 
Given the past observations $o^{(l)}_{1\to {S}}$ and $o^{(c)}_{1\to {S}}$ with length $S$, we train our model to predict $o^{(l)}_{{S+1}\to {S+T}}$ and $o^{(c)}_{{S+1}\to {S+T}}$ corresponding to future $T$ frames. The loss can be calculated by averaging joint training loss $\mathcal{L}$ on all $S+T$ frames.
We extend the generator input to the concatenation of ground truth and noisy image latent, $x^{(c)}_{in}=(z^{(c)}_{r}(\epsilon), z^{(c)}\circ m, m)$, where $r$ denotes the step to add noise, $x^{(c)}_{in}$ is the input to 2D U-Net and $m$ is a binary mask with length $S+T$ to mask out the ground truth latent for last $T$ frames. Here we ignore the time index for simplicity. On the 3D side, we directly replace the mask tokens with ground truth to enable the prediction task.

\noindent\textbf{Multi-task training policy.} 
Our training recipe is similar to the recent generative model~\citep{chen2024pixart}, which means we pretrain our model on a unimodal task and then fine-tune it on the joint training task. During the joint training stage, the model is forced to make use of both layout conditions (e.g. 3D box condition) and interaction conditions, whereas the former is fully pre-trained in the earlier stage. 
To solve this problem, we propose conditional dropping on the joint-training stage. In detail, we randomly inhibit layout conditions on only one modal. As the condition only comes from one modal, the model is naturally enforced to do cross-modal learning. Another important influence factor to our progressive training is the gap between unimodal training and joint training. We find a simple dropping strategy on the interaction is helpful enough, which means the joint training stage may go back unimodal training stage at a certain rate. Embedded with the above two policies, our joint training stage can be viewed as doing multi-task learning and on the experiment section, we show that this is important to joint training on video generation.

%% file: sec/4_exp.tex
\section{Experiments}

\subsection{Settings}

\noindent\textbf{Dataset.}
Our experiments are on the NuScenes~\citep{caesar2020nuscenes} dataset since it contains both multi-view images, lidar points, scene description text, annotations about the boxes and map.
It contains 700 videos for training and 150 videos for validation, and each video is about 20 seconds and includes about 40 key frames.
Each key frame consists of 6 camera images captured by the surround-view cameras and a point cloud captured by the LiDAR.
10 commonly used classes of 3D objects in the nuScenes following the BEVFormer~\citep{li2022bevformer}, encoded as different colors, and projected to the image space.

\noindent\textbf{Baseline methods.} 
We employ baselines for the multi-view image generation and LiDAR point cloud generation tasks, respectively.
For image generation, we compare with existing multi-view image generation methods on autonomous driving scenarios.
For LiDAR, we reproduce UltraLiDAR~\citep{xiong2023ultralidar} and use it as the baseline.

\begin{figure}[bp]
  \centering
  \includegraphics[width=\linewidth]{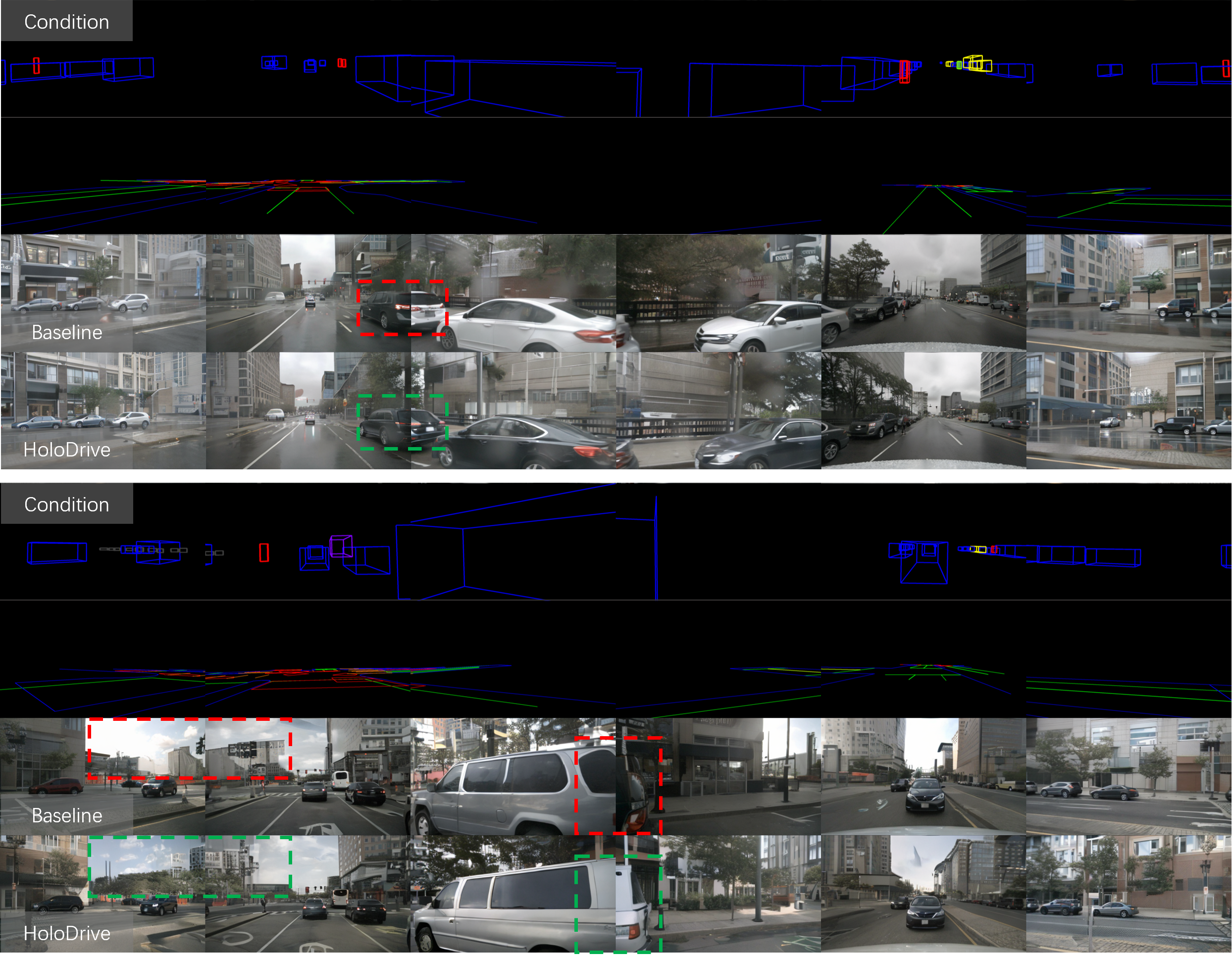}
  \captionsetup{justification=raggedright,singlelinecheck=false}
  \caption{The qualitative comparisons to the baseline method of the image generation.}
  \label{fig:image-qual-compare}
\end{figure}

\noindent\textbf{Training scheme.}
We have 3 stages of training.
The first stage trains a cross-view camera generation model starting from the SD 2.1, with newly added modules about cross-view, image condition, and depth estimation.
The second stage trains a LiDAR generation model from scratch.
The third stage trains the joint generation model starting from the previous 2 stages.
The experiments of first 2 stages are conducted on 16 V100 (32G) GPUs, and the last stage on 8 A800 (80GB) GPUs.
Images are resized to 448x256 without changing the aspect ratio largely.
The LiDAR points are clamped to the range of 100m x 100m.
For the predictive model, we use a clip of length 8, and the number of past observations is 4.
The rate of condition dropping and joint dropping are all set to 30\%.

\noindent\textbf{Evaluation metrics.}
Generated images and videos are evaluated with Frechet Inception Distance (FID)~\citep{HeuselRUNKH17} and Frechet Video Distance (FVD)~\citep{unterthiner2018towards}.
We utilize the mAP (Mean Average Precision) metric to measure the accuracy of generation, by comparing the GT location and detected location of generated results, and choose BEVFusion~\citep{liu2022bevfusion} or BEVFormer~\citep{li2022bevformer} as the detection model according to the evaluation rules of the baseline method.
Generated LiDAR points are evaluated with the Chamfer distance, L1 error (L1 Mean / Median), and relative L1 error (AbsRel Mean / Median) following the practice of 4D-Occ~\citep{khurana20234docc}.

\subsection{Main Results}

\begin{figure*}[t]
  \centering
  \includegraphics[width=\textwidth]{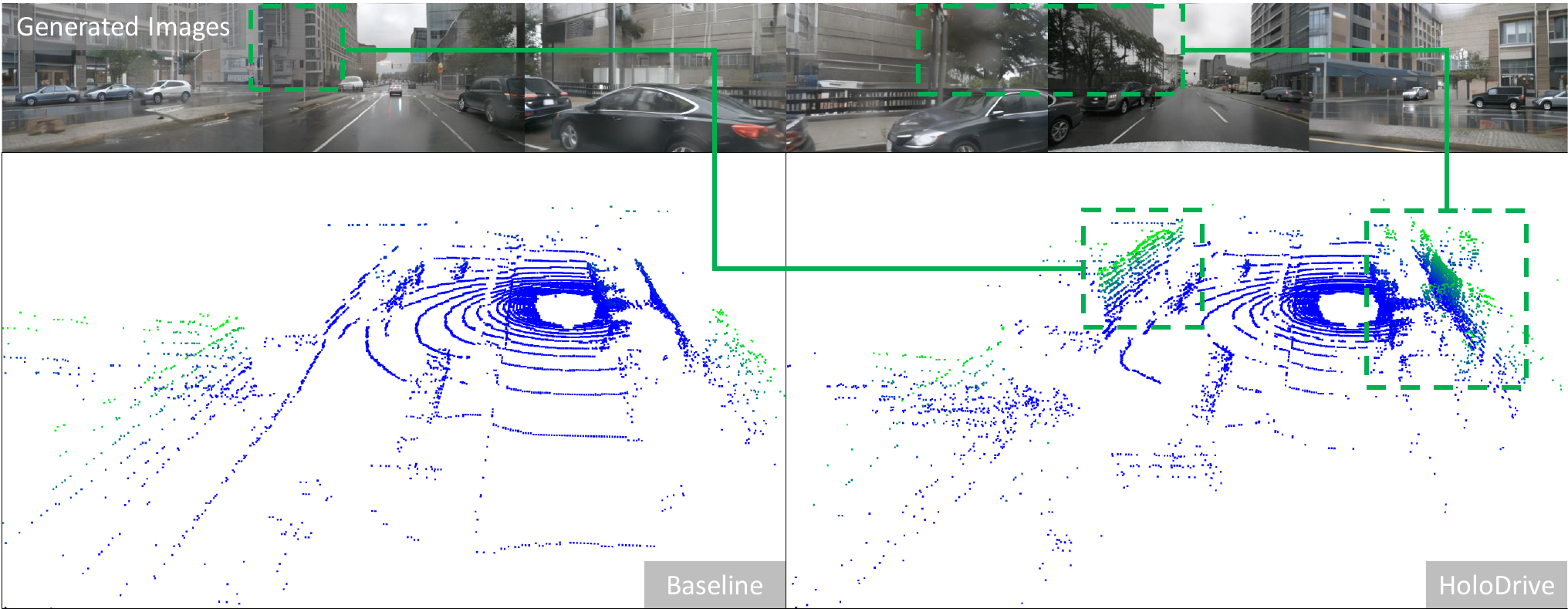}
  \caption{The qualitative comparisons to the baseline method of the LiDAR generation.}
  \label{fig:lidar-qual-compare}
\end{figure*}

\begin{figure}[bp]
  \centering
  \includegraphics[width=\linewidth]{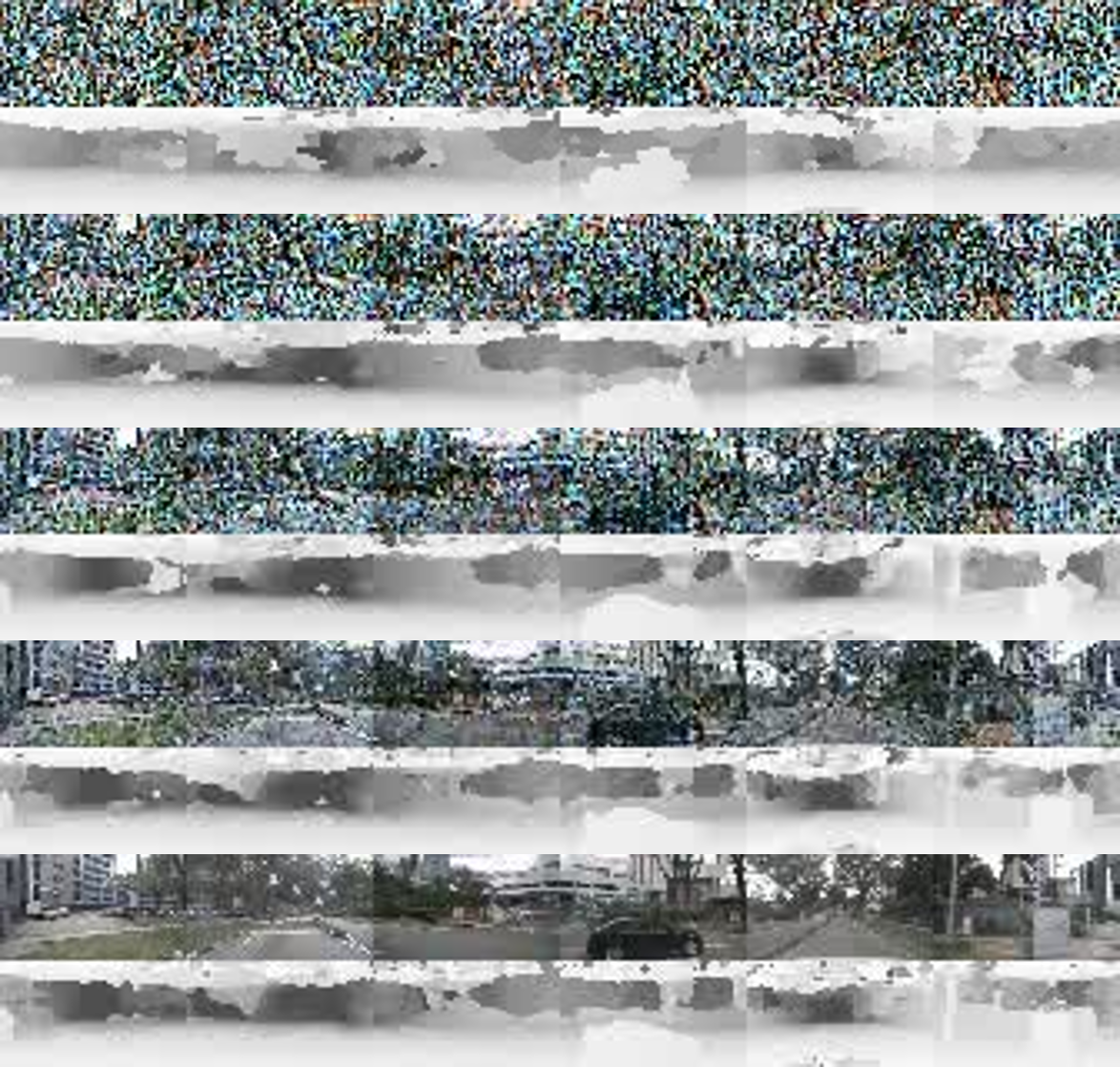}
  \caption{The estimated depth in the denoising process.}
  \label{fig:depth-visualization}
\end{figure}

\noindent\textbf{Depth estimation for image generation.}
Depth is key to cross-modal information transformation between images and point clouds.
\Cref{fig:depth-visualization} demonstrates the depth estimation capabilities of the Diffusion U-Net used as a backbone.

\noindent\textbf{Multi-view image generation.}
We compare our method with other multi-view image generation methods including the SOTA Drive-WM~\citep{wang2024driving}, and find that our HoloDrive exhibits the highest realism among all baseline methods, and is second only to Drive-WM in terms of accuracy.
The results of FID and mAPs are shown in \Cref{tab:image-generation-results}.
Qualitative results are illustrated in the \Cref{fig:image-qual-compare}.

\noindent\textbf{Single-frame LiDAR generation.}
Table~\ref{tab:lidar-gen-comp} shows the quantitative comparison with the state-of-the-art LiDAR generation method UltraLiDAR~\citep{xiong2023ultralidar}.
We reproduced the unconditional and conditional versions according to the details of the original paper.
We reported the results of two types of our method: 2d→3d and 2d$\leftrightarrow$3d (2D-3D joint-training).
3D condition (3D boxes and HDMap) improves all the scores of the LiDAR quality.
Incorporating 2D features from 2D model into our 3D models significantly enhances Chamfer, L1 error, and AbsRel.
Finally, with the interaction between 2D and 3D models, our method shows better LiDAR generation quality, as details of trees and buildings in the point cloud generated in the example shown in \Cref{fig:lidar-qual-compare}.

\noindent\textbf{Cross-modal consistency.}
One clear advantage of our proposed joint 2D-3D generation approach is cross-modal consistency.
As presented in \Cref{fig:visualization}, the generated 2D multi-view street scenes are highly consistent with the 3D LiDAR points, probably owing to the frequent interactions between the two modalities during training and inference.

\noindent\textbf{LiDAR prediction.}
We follow the implement details of Copilot4D~\citep{zhang2023copilot4d} to construct our 3D world model. It is worth noting that we set the ego car as the coordinate origin during sequence generation, rather than fixing it to one reference frame. The results are shown in Table \ref{tab:lidar-pred-comp}, our re-implementation achieves similar results compared to Copilot4D and outperforms previous methods. 

\begin{table*}[t]
  \centering
  \begin{tabular}{@{\hspace{0.3cm}}c@{\hspace{0.6cm}}c@{\hspace{0.3cm}}ccccc|cccc}
  \toprule
  \multicolumn{3}{c}{\textbf{Trainable}} & \multirow{2}{*}{\textbf{3D$\to$2D}} & \multirow{2}{*}{\textbf{2D$\to$3D}} & \multicolumn{2}{c|}{\textbf{Drop ratio}} & \multirow{2}{*}{\textbf{FID$\downarrow$}} & \multirow{2}{*}{\textbf{FVD$\downarrow$}} & \multirow{2}{*}{\textbf{Chamfer$\downarrow$}} & \textbf{L1} \\
  \cline{0-2} \cline{6-7}
  2D & 3D & Depth &  & & Condition & Joint & & & & \textbf{Mean$\downarrow$} \\
  \midrule
  $\surd$ & & & & & - & - & 12.7 & 136 & - & -\\
  & $\surd$ & & & & - & - & - & - & 0.890 & 1.532\\
  $\surd$ & & $\surd$ & & & 0\% & 0\% & 11.6 & 140 & - & - \\
  $\surd$ & $\surd$ & $\surd$ & $\surd$ & & 0\% & 0\% & 11.3 & 126 & 0.891 & 1.508 \\
  $\surd$ & $\surd$ & $\surd$ & $\surd$ & $\surd$ & 0\% & 0\% & 11.6 & 117 & 0.901 & 1.499 \\
  $\surd$ & $\surd$ & $\surd$ & $\surd$ & $\surd$ & 30\% & 0\% & 10.7 & 120 & 0.849 & 1.490 \\
  $\surd$ & $\surd$ & $\surd$ & $\surd$ & $\surd$ & 30\% & 30\% & \textbf{9.4} & \textbf{83} & \textbf{0.838} & \textbf{1.469}\\
  \bottomrule
  \end{tabular}
  \label{tab:video-abl}
  \caption{Ablations on temporal joint training. All metrics are evaluated on 8 frames.}
\end{table*}

\begin{table}[bp]
  \centering
  \begin{tabular}{c|ccc}
  \toprule
  \textbf{Method} & \textbf{Chamfer$\downarrow$} & \textbf{\makecell[c]{L1\\Med$\downarrow$}} & \textbf{\makecell[c]{L1\\Mean$\downarrow$}} \\
  \midrule
  S2Net \cite{weng2022s2net} & 2.06 & - & - \\
  Occ4D \cite{khurana20234docc} & 1.40 & 0.43 & - \\
  Copilot4D \cite{zhang2023copilot4d} & 1.40 & \textbf{0.13} & \textbf{1.23} \\
  \midrule
  HoloDrive (3D) & 0.89 & \textbf{0.13} & 1.53 \\
  HoloDrive (Joint) & \textbf{0.83} & \textbf{0.13} & 1.46 \\
  \bottomrule
  \end{tabular}
  \caption{LiDAR prediction comparison.}
  \label{tab:lidar-pred-comp}
\end{table}

\noindent\textbf{Predictive world model.}
We further make a comparison with other methods. We follow the evaluation pipeline in Drive-WM~\citep{wang2024driving}. Especially, for each validation video in NuScenes, we generate corresponding 40 frames in an auto-regressive manner~\citep{blattmann2023videoldm} and pick 16 frames for evaluation. The results are shown in Table \ref{tab:video-main}. Our method outperforms previous methods other than FVD on DriveDreamer-2~\citep{zhao2024drivedreamer} and some of the reason lies in the usage of SVD: the ablation study in~\citep{zhao2024drivedreamer} showed that simply changing SD1.5 to SVD can significantly reduce FVD from 340.8 to 94.6.

\subsection{Ablation Study on Video and LiDAR Joint Generation}
We conduct ablation to test different training designs on the predictive world model.
For efficiency, we use only one auto-regressive step, which generates a total of 8 frames.
We examine the effect of depth supervision, which achieves FID 11.6 and FVD 140, showing better results compared with the 2D baseline.
The reason may be that depth supervision enables the generation model to perceive the scene structure, thereby improving image generation.
The cross-modal interaction only leads to minor improvement without introducing dropping policies.
As our model is pre-trained without joint interaction, the model focuses on single-modal generation, whose ability may struggle to directly learn to use both layout conditions and cross-modal interaction.

To solve this problem, we propose two types of dropping on the video domain, which randomly drops the condition (e.g. past observations) and the 2D-3D interaction.
These policies can be seen as a clearly defined multi-task framework with both single-modal and cross-modal generation.
The final result with both condition and joint dropping verifies our hypothesis, achieving consistent improvement on all metrics.

%% file: sec/5_conclusion.tex
\section{Conclusion}

In this work, we propose a novel framework, namely \ourmethod, for 2D-3D joint generation on multi-view camera images and LiDAR point clouds.
We perform joint generation via building transform modules between Camera and BEV spaces, with the help of additional depth supervision.
We apply our joint pipeline to both single-frame generation and video prediction with carefully designed progressive training stages.
We conduct extensive experiments on single frame generation and world model benchmarks on the NuScence dataset.
Compared with state-of-the-art methods, our proposed \ourmethod~achieves significant improvements in terms of generation metrics.

%% file: sec/X_suppl.tex
\clearpage
\setcounter{page}{1}
\maketitlesupplementary

\section{Discussions}
\noindent\textbf{Why use MaskGit on 3D branch?}
Our decision to utilize this architecture is inspired by an existing approach, Copilot4D, which has already demonstrated its effectiveness in LiDAR generation. The established metric values from this pre-existing approach further aid us in evaluating the quality of our generated LiDAR results. Moreover, the core of joint generation come from the feature interaction and time-step alignment, so our method can be extended to other point cloud generation methods in the future. 

\noindent\textbf{Difference with BEVWorld.}
Our approach is inspired by fostering an information exchange between 2D and 3D generation models through a modular plugin design. This strategy yields a key benefit: our results either match or surpass the quality of the original 2D or 3D modules. Nevertheless, in the case of BEVWorld, which necessitates training a 2D-3D auto-encoder from the ground up, the generation quality is compromised due to the overlooking of well-established 2D generation models.

\section{More Experiment Details}

\noindent\textbf{Depth estimation net.}
In our experiment, we use $D=96$, and the actual distance for each depth bin ranging from 1.0 m to 58.6 m.

\noindent\textbf{3D $\to$ 2D details.}
We sample the output features of the 1st and 2nd down-block of the LiDAR generation model to calculate the $e^{(\texttt{proj})}$.

\noindent\textbf{2D $\to$ 3D details.}
We sample the output features of the 1st and 2nd down-block of the image generation model to calculate the $e^{(\texttt{proj})}$.

\noindent\textbf{Training the image generation model.}
The model is initialized from the SD 2.1 checkpoint and the additional depth estimation and condition adapter nets are initialized from scratch.
The condition features are added to the backbone block features by zero initialized convolution modules.
We use the AdamW optimizer with learning rate $6\times 10^{-5}$. The model is fine-tuned on the nuScenes~\cite{caesar2020nuscenes} images and conditions for 20,000 steps with a total batch size of 64.

\noindent\textbf{Train the LiDAR generation model.}
Conditional LiDAR generation model utilizes a pre-trained and frozen ResNet-50 \cite{he2016deep} to extract the spatial features of 3D box and HDMap conditions.
We use the AdamW optimizer and learning rate $4\times 10^{-4}$. The model is trained from scratch on the nuScenes~\cite{caesar2020nuscenes} LiDAR point cloud for 18,000 steps with a total batch size of 256.

\noindent\textbf{Train the joint generation model.}
The 2D-3D joint generation model loads the pre-trained image generation model, LiDAR generation model from the previous training stages.
The output features of cross-view module are mixed with original features with learnable parameters initialized as a small ratio, to prevent a sudden transition at the early phase of joint training.
We use AdamW optimizer with learning rate $5\times 10^{-5}$ for this stage. The model is trained for 30,000 steps with the total batch size of 192.

\noindent\textbf{Details of temporal blocks.} 
Two kinds of temporal blocks are applied in our model: temporal attention block and temporal residual block.
We use GPT-2 style temporal attention blocks \cite{radford2019language} to attend over the same feature location across time and for the temporal residual block, simple 3D convolution and residual connection is used.
In our 2D model, we append one temporal attention or residual block to each spatial attention or residual block.
To align with Copilot4D~\cite{zhang2023copilot4d}, we only add one temporal attention block after two spatial attention blocks.

\noindent\textbf{Training the temporal joint generation model.}
The training pipeline of temporal model share the same idea with spatial counterpart.
We first pretrain unimodal generation models for both 2D and 3D and then jointly fine-tune them.
We directly load the parameters of our image generation model and train 30,000 steps with temporal loss.
Similarly, we train the 3D temporal model loaded from single frame model with 80,000 steps.
Finally, we tune the temporal joint generation model with 30,000 steps.

\begin{table}[h]
  \centering
  \begin{tabular}{ccc|cc}
  \toprule
  \textbf{2D condition} & \textbf{3D$\to$2D} & \textbf{3D$\leftrightarrow$2D} & \textbf{FID$\downarrow$} & \textbf{mAP$\uparrow$} \\
  \toprule
  & & & 54.63 & - \\
  $\surd$ & & & 14.41 & - \\
  \midrule
  $\surd$ & $\surd$ & & 11.87 & 18.64 \\
  $\surd$ & $\surd$ & $\surd$ & \textbf{10.64} & \textbf{19.98} \\
  \bottomrule
  \end{tabular}
  \caption{Ablation of image generation}
  \label{tab:image-generation-ablation}
\end{table}

\begin{table}[h]
  \centering
  \begin{tabular}{cc|ccc}
  \toprule
  \textbf{2D$\rightarrow$3D} & \textbf{2D$\leftrightarrow$3D} & \textbf{Chamfer$\downarrow$} & \textbf{\makecell[c]{L1\\Mean$\downarrow$}} & \textbf{\makecell[c]{AbsRel L1\\Mean(\%)$\downarrow$}} \\
  \toprule
  & & 10.37 & 3.98 & 35.14 \\
  $\surd$ & & 7.62 & 2.38 & 17.17 \\
  $\surd$ & $\surd$ & \textbf{7.61} & \textbf{2.37} & \textbf{16.89} \\
  \bottomrule
  \end{tabular}
  \caption{Ablations on LiDAR generation}
  \label{tab:lidar-abl}
\end{table}

\section{More Ablation Studies}

\begin{figure*}[t]
  \centering
  \includegraphics[width=\textwidth]{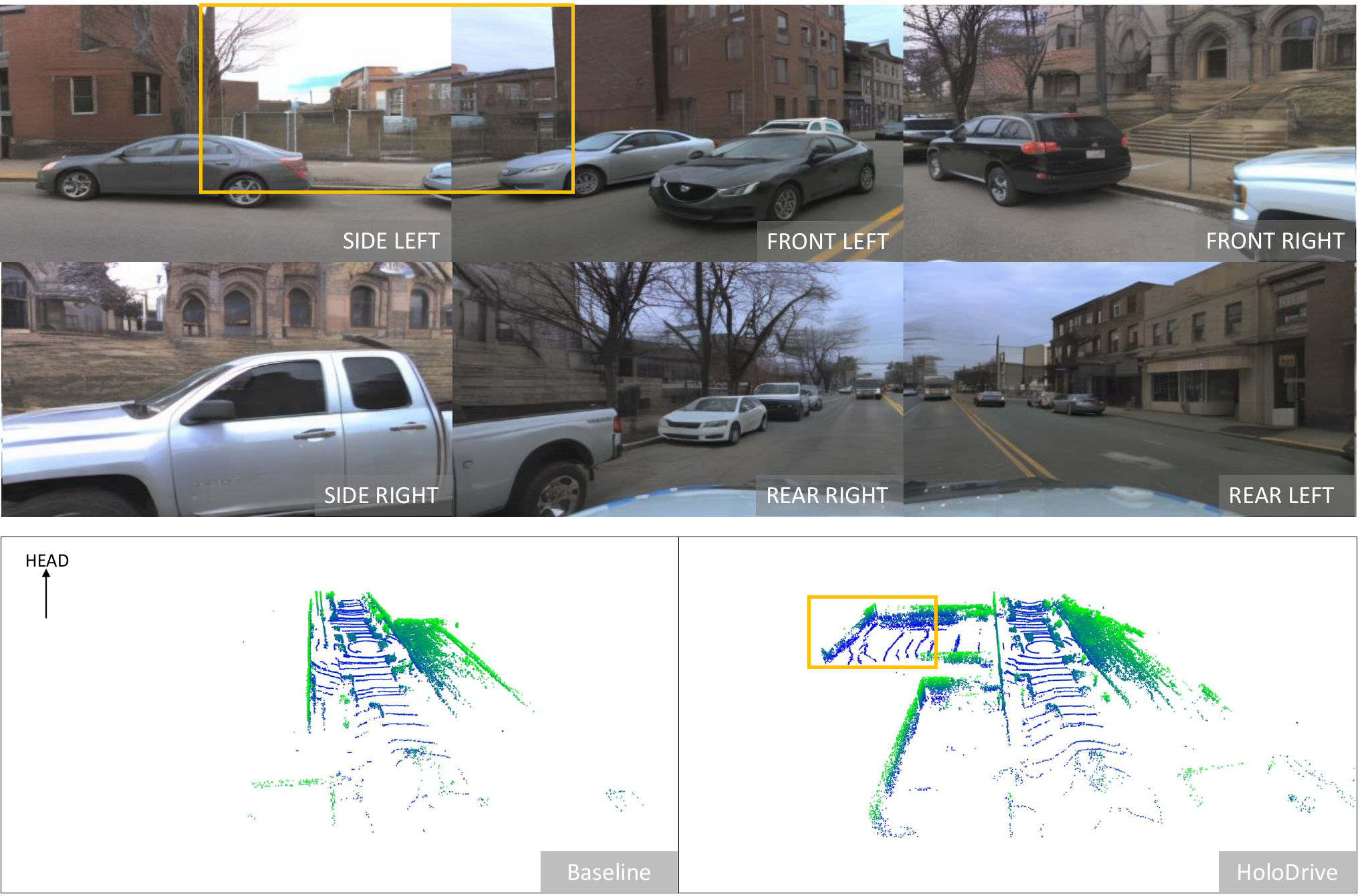}
  \caption{Examples on the Argoverse datset.}
  \label{fig:result-argo-1}
\end{figure*}

\begin{figure*}[bp]
  \centering
  \includegraphics[width=\textwidth]{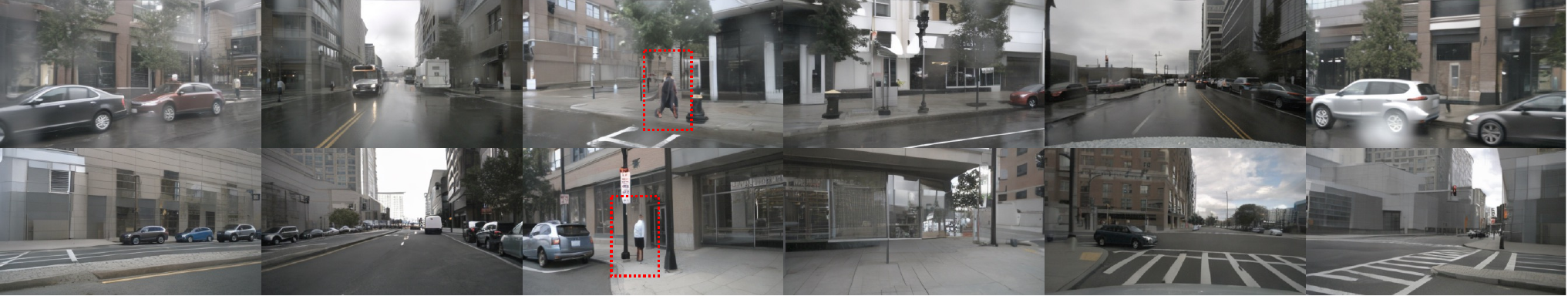}
  \caption{Distorted Pedestrians.}
  \label{fig:failure_case_ped}
\end{figure*}

\noindent\textbf{Image generation.}
In \Cref{tab:image-generation-ablation}, we explore how different setups affect the realism and accuracy of the generated images.
The model that produces the results in the 1st row is trained with nuScenes images, and the model that produces the results in the 2nd row is trained with both image and conditions. Both first 2 rows are not multi-view image generation, so we do not evaluate mAP on them.
The model that produces the results in the 3rd row is trained with the features from LiDAR generation model by 3D-to-2D module. Under this setting, both image and LiDAR generation models are trainable with their own training objectives.
The model that produces the results in the 4th row is jointly trained with bi-directional feature interaction between image and LiDAR parts.

\noindent\textbf{LiDAR generation.} 
We then conduct ablation study on LiDAR generation to investigate the mechanism that influences the LiDAR generation quality. In Table~\ref{tab:lidar-abl}, we report the Chamfer distance within 50 m, L1 Mean, and AbsRel. Using the 3D condition reduces Chamfer by about 6, L1 average by about 0.5, and AbsRel by about 16\%, respectively. Integrating
2D features and keeping the 2D model frozen improve the L1 Mean and AbsRel, which validates the effectiveness of multi-view 2D features for the 3D generation. Fine-tuning the 2D model or using the 2D loss also continuously improves the LiDAR generation quality and achieves the best Chamfer. Finally, with the interaction between 2D and 3D models, our method achieved the best L1 Mean of 2.55, and AbsRel of 19.37, though Chamfer drops a little, which leave for future investigation.

\section{More Qualitative Results}

\noindent\textbf{Results on the Argoverse.}
We conducted experiments on the Argoverse~\citep{NEURIPSDATASETSANDBENCHMARKS2021_4734ba6f} dataset and verified the generalizability of the method.
Fig.~\ref{fig:result-argo-1} illustrates that HoloDrive can generate high-quality images and point clouds with considerable cross-modal consistency on Argoverse dataset. we use yellow box to specify consistent results missed on baseline.

\begin{figure*}[t]
  \centering
  \includegraphics[width=\textwidth]{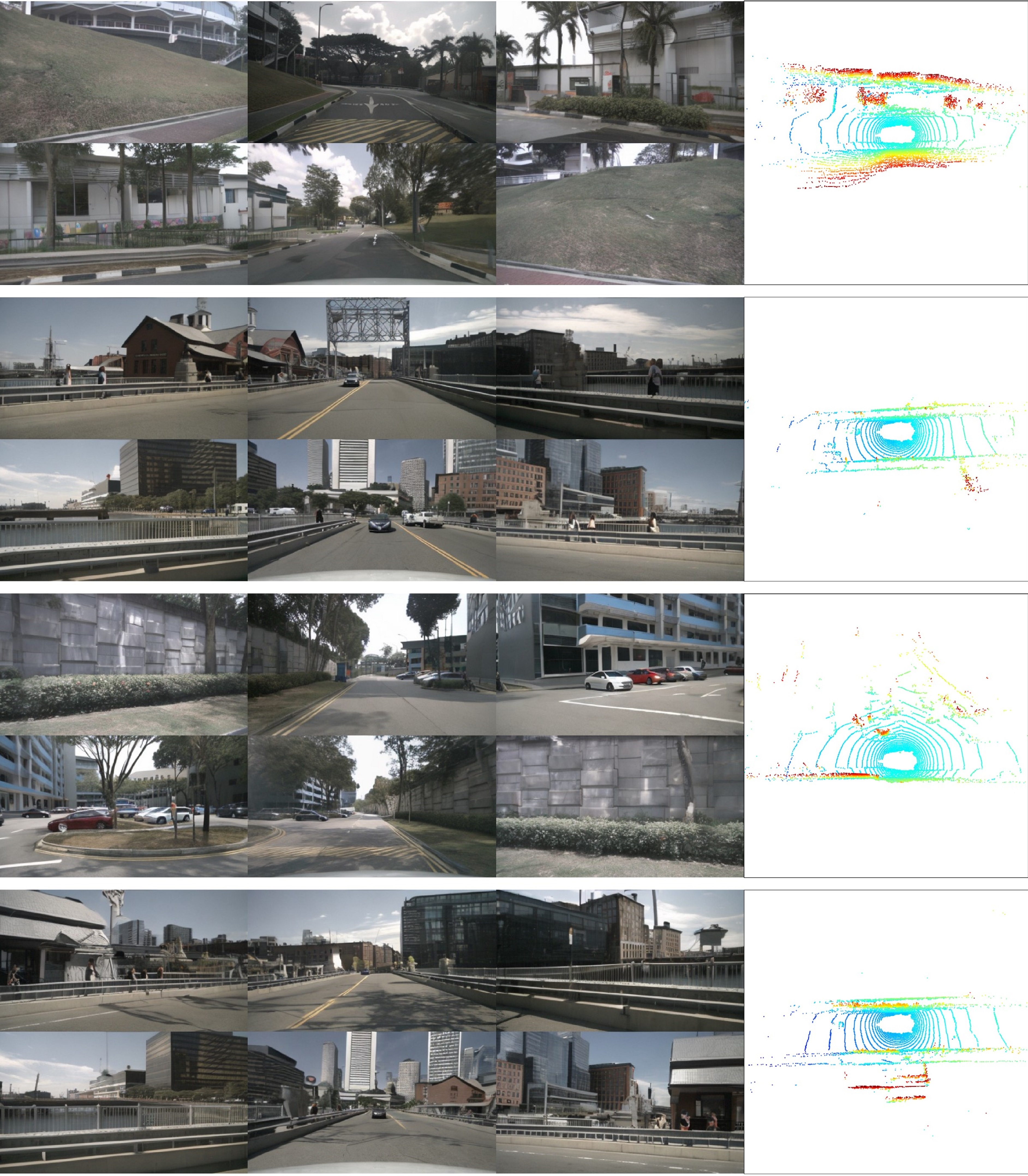}
  \caption{2D-3D consistent generation from HoloDrive.}
  \label{fig:qual-result-1}
\end{figure*}

\begin{figure*}[t]
  \centering
  \includegraphics[width=\textwidth]{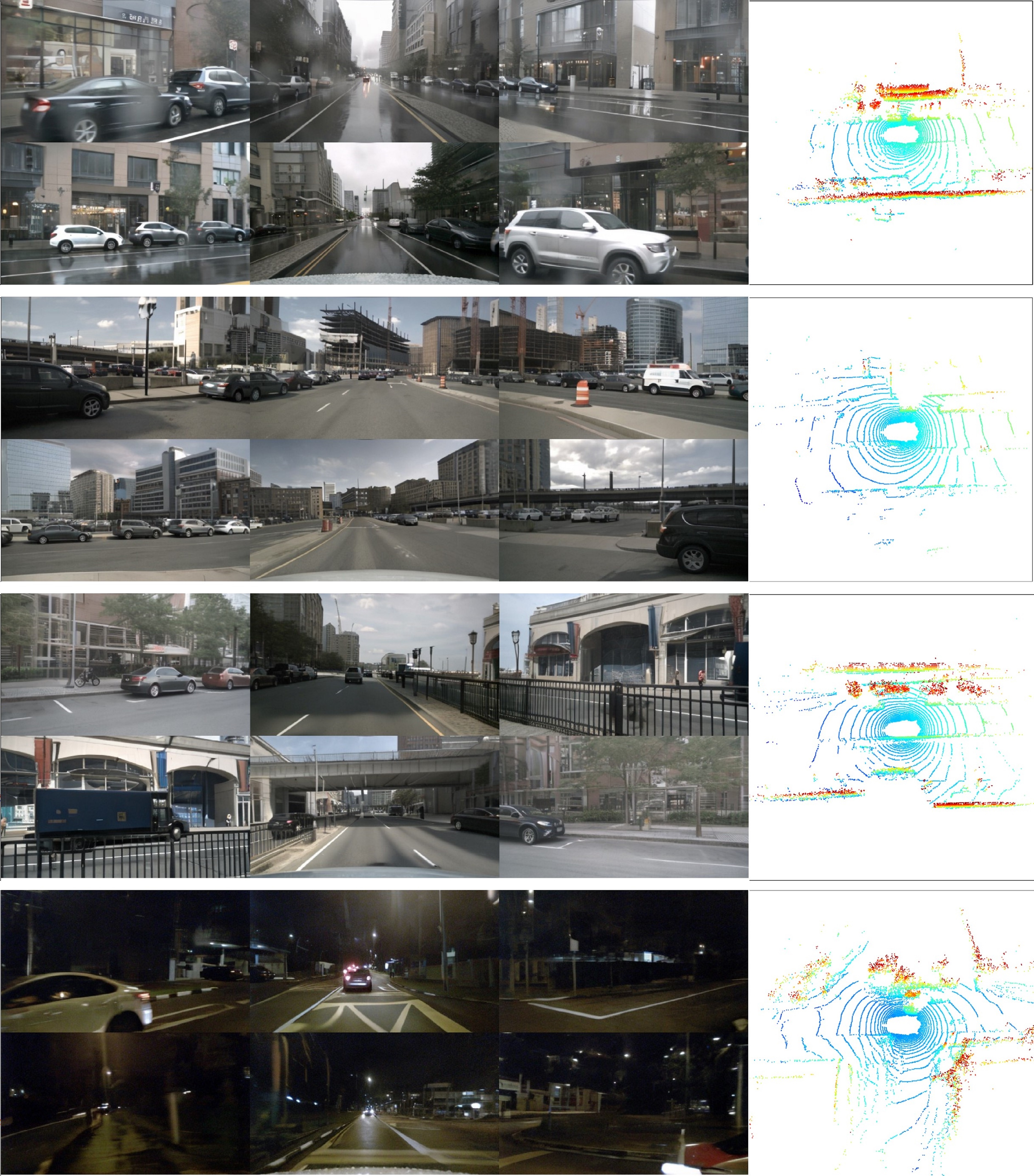}
  \caption{2D-3D consistent generation from HoloDrive.}
  \label{fig:qual-result-2}
\end{figure*}

\noindent\textbf{2D-3D consistency.}
Fig.~\ref{fig:qual-result-1} and Fig.~\ref{fig:qual-result-2} illustrate that HoloDrive can generate high-quality images and point clouds with considerable cross-modal consistency.
This enables applications as a multi-modality neural simulator.

\begin{figure*}[ht]
  \centering
  \includegraphics[width=\textwidth]{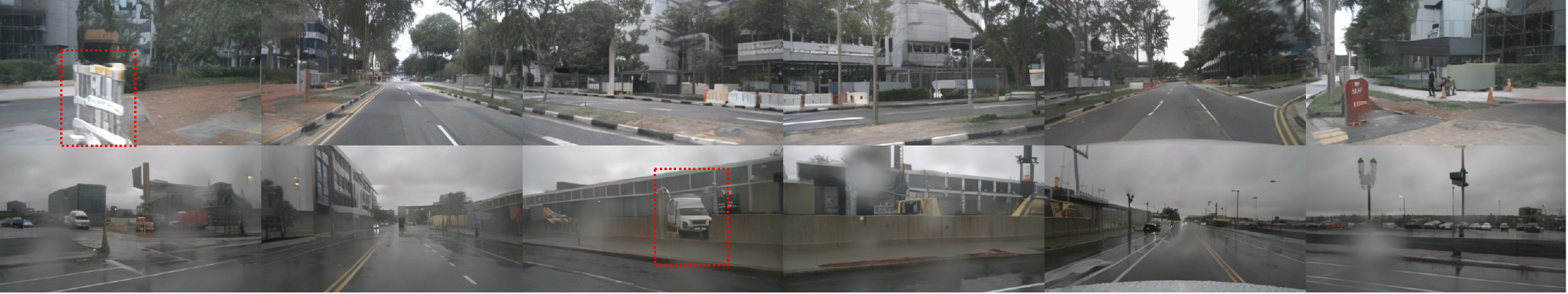}
  \caption{Unreasonable Elements.}
  \label{fig:failure_case_truck}
\end{figure*}

\noindent\textbf{Failure cases.}
Although HoloDrive can generate realistic street view scenarios, it still has some limitations.
Similar to other street view generation methods, the predicted images often contain distorted pedestrians (as in Fig.~\ref{fig:failure_case_ped}).
Extra refinement for pedestrians could potentially beneficial. There are still some unreasonable elements in generation results as Fig.~\ref{fig:failure_case_truck} reveals.
Incorporating more datasets could be helpful.

%% file: main.bbl
\begin{thebibliography}{54}
\providecommand{\natexlab}[1]{#1}
\providecommand{\url}[1]{\texttt{#1}}
\expandafter\ifx\csname urlstyle\endcsname\relax
  \providecommand{\doi}[1]{doi: #1}\else
  \providecommand{\doi}{doi: \begingroup \urlstyle{rm}\Url}\fi

\bibitem[Blattmann et~al.(2023)Blattmann, Rombach, Ling, Dockhorn, Kim, Fidler, and Kreis]{blattmann2023videoldm}
Andreas Blattmann, Robin Rombach, Huan Ling, Tim Dockhorn, Seung~Wook Kim, Sanja Fidler, and Karsten Kreis.
\newblock Align your latents: High-resolution video synthesis with latent diffusion models.
\newblock In \emph{IEEE Conference on Computer Vision and Pattern Recognition ({CVPR})}, 2023.

\bibitem[Caccia et~al.(2019)Caccia, Van~Hoof, Courville, and Pineau]{caccia2019lidarvae}
Lucas Caccia, Herke Van~Hoof, Aaron Courville, and Joelle Pineau.
\newblock Deep generative modeling of lidar data.
\newblock In \emph{2019 IEEE/RSJ International Conference on Intelligent Robots and Systems (IROS)}, pages 5034--5040. IEEE, 2019.

\bibitem[Caesar et~al.(2020)Caesar, Bankiti, Lang, Vora, Liong, Xu, Krishnan, Pan, Baldan, and Beijbom]{caesar2020nuscenes}
Holger Caesar, Varun Bankiti, Alex~H Lang, Sourabh Vora, Venice~Erin Liong, Qiang Xu, Anush Krishnan, Yu Pan, Giancarlo Baldan, and Oscar Beijbom.
\newblock nuscenes: A multimodal dataset for autonomous driving.
\newblock In \emph{Proceedings of the IEEE/CVF conference on computer vision and pattern recognition}, pages 11621--11631, 2020.

\bibitem[Chang et~al.(2022)Chang, Zhang, Jiang, Liu, and Freeman]{chang2022maskgit}
Huiwen Chang, Han Zhang, Lu Jiang, Ce Liu, and William~T Freeman.
\newblock Maskgit: Masked generative image transformer.
\newblock In \emph{Proceedings of the IEEE/CVF Conference on Computer Vision and Pattern Recognition}, pages 11315--11325, 2022.

\bibitem[Chen et~al.(2024)Chen, Ge, Xie, Wu, Yao, Ren, Wang, Luo, Lu, and Li]{chen2024pixart}
Junsong Chen, Chongjian Ge, Enze Xie, Yue Wu, Lewei Yao, Xiaozhe Ren, Zhongdao Wang, Ping Luo, Huchuan Lu, and Zhenguo Li.
\newblock Pixart-$\backslash$sigma: Weak-to-strong training of diffusion transformer for 4k text-to-image generation.
\newblock \emph{arXiv preprint arXiv:2403.04692}, 2024.

\bibitem[Gao et~al.(2024)Gao, Chen, Xie, Lanqing, Li, Yeung, and Xu]{gao2023magicdrive}
Ruiyuan Gao, Kai Chen, Enze Xie, HONG Lanqing, Zhenguo Li, Dit-Yan Yeung, and Qiang Xu.
\newblock Magicdrive: Street view generation with diverse 3d geometry control.
\newblock In \emph{The Twelfth International Conference on Learning Representations}, 2024.

\bibitem[Goodfellow et~al.(2014)Goodfellow, Pouget-Abadie, Mirza, Xu, Warde-Farley, Ozair, Courville, and Bengio]{goodfellow2014generative}
Ian Goodfellow, Jean Pouget-Abadie, Mehdi Mirza, Bing Xu, David Warde-Farley, Sherjil Ozair, Aaron Courville, and Yoshua Bengio.
\newblock Generative adversarial nets.
\newblock \emph{Advances in neural information processing systems}, 27, 2014.

\bibitem[He et~al.(2016)He, Zhang, Ren, and Sun]{he2016deep}
Kaiming He, Xiangyu Zhang, Shaoqing Ren, and Jian Sun.
\newblock Deep residual learning for image recognition.
\newblock In \emph{Proceedings of the IEEE conference on computer vision and pattern recognition}, pages 770--778, 2016.

\bibitem[Heusel et~al.(2017)Heusel, Ramsauer, Unterthiner, Nessler, and Hochreiter]{HeuselRUNKH17}
Martin Heusel, Hubert Ramsauer, Thomas Unterthiner, Bernhard Nessler, and Sepp Hochreiter.
\newblock Gans trained by a two time-scale update rule converge to a local nash equilibrium.
\newblock \emph{Advances in neural information processing systems}, 30, 2017.

\bibitem[Ho and Salimans(2021)]{ho2021classifier}
Jonathan Ho and Tim Salimans.
\newblock Classifier-free diffusion guidance.
\newblock In \emph{NeurIPS 2021 Workshop on Deep Generative Models and Downstream Applications}, 2021.

\bibitem[Hu et~al.(2023)Hu, Russell, Yeo, Murez, Fedoseev, Kendall, Shotton, and Corrado]{hu2023gaia}
Anthony Hu, Lloyd Russell, Hudson Yeo, Zak Murez, George Fedoseev, Alex Kendall, Jamie Shotton, and Gianluca Corrado.
\newblock Gaia-1: A generative world model for autonomous driving.
\newblock \emph{arXiv preprint arXiv:2309.17080}, 2023.

\bibitem[Hu et~al.(2024)Hu, Zhang, and Hu]{hu2024rangeldm}
Qianjiang Hu, Zhimin Zhang, and Wei Hu.
\newblock Rangeldm: Fast realistic lidar point cloud generation.
\newblock \emph{arXiv preprint arXiv:2403.10094}, 2024.

\bibitem[Huang et~al.(2024)Huang, Wen, Zhao, Hu, Liu, Jia, Mao, Wang, Zhang, Chen, et~al.]{huang2024subjectdrive}
Binyuan Huang, Yuqing Wen, Yucheng Zhao, Yaosi Hu, Yingfei Liu, Fan Jia, Weixin Mao, Tiancai Wang, Chi Zhang, Chang~Wen Chen, et~al.
\newblock Subjectdrive: Scaling generative data in autonomous driving via subject control.
\newblock \emph{arXiv preprint arXiv:2403.19438}, 2024.

\bibitem[Jia et~al.(2023)Jia, Mao, Liu, Zhao, Wen, Zhang, Zhang, and Wang]{jia2023adriver}
Fan Jia, Weixin Mao, Yingfei Liu, Yucheng Zhao, Yuqing Wen, Chi Zhang, Xiangyu Zhang, and Tiancai Wang.
\newblock Adriver-i: A general world model for autonomous driving.
\newblock \emph{arXiv preprint arXiv:2311.13549}, 2023.

\bibitem[Karras et~al.(2022)Karras, Aittala, Aila, and Laine]{karras2022elucidating}
Tero Karras, Miika Aittala, Timo Aila, and Samuli Laine.
\newblock Elucidating the design space of diffusion-based generative models.
\newblock \emph{Advances in Neural Information Processing Systems}, 35:\penalty0 26565--26577, 2022.

\bibitem[Khurana et~al.(2023)Khurana, Hu, Held, and Ramanan]{khurana20234docc}
Tarasha Khurana, Peiyun Hu, David Held, and Deva Ramanan.
\newblock Point cloud forecasting as a proxy for 4d occupancy forecasting.
\newblock In \emph{Proceedings of the IEEE/CVF Conference on Computer Vision and Pattern Recognition}, pages 1116--1124, 2023.

\bibitem[Kim et~al.(2021)Kim, Philion, Torralba, and Fidler]{kim2021drivegan}
Seung~Wook Kim, Jonah Philion, Antonio Torralba, and Sanja Fidler.
\newblock Drivegan: Towards a controllable high-quality neural simulation.
\newblock In \emph{Proceedings of the IEEE/CVF Conference on Computer Vision and Pattern Recognition}, pages 5820--5829, 2021.

\bibitem[Kingma and Welling(2013)]{kingma2013auto}
Diederik~P Kingma and Max Welling.
\newblock Auto-encoding variational bayes.
\newblock \emph{arXiv preprint arXiv:1312.6114}, 2013.

\bibitem[Li et~al.(2022{\natexlab{a}})Li, Ge, Yu, Yang, Wang, Shi, Sun, and Li]{li2022bevdepth}
Yinhao Li, Zheng Ge, Guanyi Yu, Jinrong Yang, Zengran Wang, Yukang Shi, Jianjian Sun, and Zeming Li.
\newblock Bevdepth: Acquisition of reliable depth for multi-view 3d object detection.
\newblock \emph{arXiv preprint arXiv:2206.10092}, 2022{\natexlab{a}}.

\bibitem[Li et~al.(2022{\natexlab{b}})Li, Yu, Meng, Caine, Ngiam, Peng, Shen, Lu, Zhou, Le, et~al.]{li2022deepfusion}
Yingwei Li, Adams~Wei Yu, Tianjian Meng, Ben Caine, Jiquan Ngiam, Daiyi Peng, Junyang Shen, Yifeng Lu, Denny Zhou, Quoc~V Le, et~al.
\newblock Deepfusion: Lidar-camera deep fusion for multi-modal 3d object detection.
\newblock In \emph{Proceedings of the IEEE/CVF Conference on Computer Vision and Pattern Recognition}, pages 17182--17191, 2022{\natexlab{b}}.

\bibitem[Li et~al.(2023)Li, Liu, Wu, Mu, Yang, Gao, Li, and Lee]{li2023gligen}
Yuheng Li, Haotian Liu, Qingyang Wu, Fangzhou Mu, Jianwei Yang, Jianfeng Gao, Chunyuan Li, and Yong~Jae Lee.
\newblock Gligen: Open-set grounded text-to-image generation.
\newblock In \emph{Proceedings of the IEEE/CVF Conference on Computer Vision and Pattern Recognition}, pages 22511--22521, 2023.

\bibitem[Li et~al.(2022{\natexlab{c}})Li, Wang, Li, Xie, Sima, Lu, Qiao, and Dai]{li2022bevformer}
Zhiqi Li, Wenhai Wang, Hongyang Li, Enze Xie, Chonghao Sima, Tong Lu, Yu Qiao, and Jifeng Dai.
\newblock Bevformer: Learning bird’s-eye-view representation from multi-camera images via spatiotemporal transformers.
\newblock \emph{arXiv preprint arXiv:2203.17270}, 2022{\natexlab{c}}.

\bibitem[Li et~al.(2024)Li, Zhang, Lin, Xiong, Long, Deng, Zhang, Liu, Huang, Xiao, et~al.]{li2024hunyuan}
Zhimin Li, Jianwei Zhang, Qin Lin, Jiangfeng Xiong, Yanxin Long, Xinchi Deng, Yingfang Zhang, Xingchao Liu, Minbin Huang, Zedong Xiao, et~al.
\newblock Hunyuan-dit: A powerful multi-resolution diffusion transformer with fine-grained chinese understanding.
\newblock \emph{arXiv preprint arXiv:2405.08748}, 2024.

\bibitem[Liang et~al.(2022)Liang, Xie, Yu, Xia, Lin, Wang, Tang, Wang, and Tang]{liang2022bevfusion}
Tingting Liang, Hongwei Xie, Kaicheng Yu, Zhongyu Xia, Zhiwei Lin, Yongtao Wang, Tao Tang, Bing Wang, and Zhi Tang.
\newblock Bevfusion: A simple and robust lidar-camera fusion framework.
\newblock \emph{Advances in Neural Information Processing Systems}, 35:\penalty0 10421--10434, 2022.

\bibitem[Liu et~al.(2021)Liu, Lin, Cao, Hu, Wei, Zhang, Lin, and Guo]{liu2021swin}
Ze Liu, Yutong Lin, Yue Cao, Han Hu, Yixuan Wei, Zheng Zhang, Stephen Lin, and Baining Guo.
\newblock Swin transformer: Hierarchical vision transformer using shifted windows.
\newblock In \emph{Proceedings of the IEEE/CVF international conference on computer vision}, pages 10012--10022, 2021.

\bibitem[Liu et~al.(2023)Liu, Tang, Amini, Yang, Mao, Rus, and Han]{liu2022bevfusion}
Zhijian Liu, Haotian Tang, Alexander Amini, Xingyu Yang, Huizi Mao, Daniela Rus, and Song Han.
\newblock Bevfusion: Multi-task multi-sensor fusion with unified bird's-eye view representation.
\newblock In \emph{IEEE International Conference on Robotics and Automation (ICRA)}, 2023.

\bibitem[Lu et~al.(2024)Lu, Wang, Huang, Wu, Liu, Ouyang, and Bai]{lu2024fit}
Zeyu Lu, Zidong Wang, Di Huang, Chengyue Wu, Xihui Liu, Wanli Ouyang, and Lei Bai.
\newblock Fit: Flexible vision transformer for diffusion model.
\newblock \emph{arXiv preprint arXiv:2402.12376}, 2024.

\bibitem[Ma et~al.(2024)Ma, Goldstein, Albergo, Boffi, Vanden-Eijnden, and Xie]{ma2024sit}
Nanye Ma, Mark Goldstein, Michael~S Albergo, Nicholas~M Boffi, Eric Vanden-Eijnden, and Saining Xie.
\newblock Sit: Exploring flow and diffusion-based generative models with scalable interpolant transformers.
\newblock \emph{arXiv preprint arXiv:2401.08740}, 2024.

\bibitem[Mersch et~al.(2022)Mersch, Chen, Behley, and Stachniss]{mersch2022ST3DCNN}
Benedikt Mersch, Xieyuanli Chen, Jens Behley, and Cyrill Stachniss.
\newblock Self-supervised point cloud prediction using 3d spatio-temporal convolutional networks.
\newblock In \emph{Conference on Robot Learning}, pages 1444--1454. PMLR, 2022.

\bibitem[Mou et~al.(2023)Mou, Wang, Xie, Wu, Zhang, Qi, Shan, and Qie]{mou2023t2i}
Chong Mou, Xintao Wang, Liangbin Xie, Yanze Wu, Jian Zhang, Zhongang Qi, Ying Shan, and Xiaohu Qie.
\newblock T2i-adapter: Learning adapters to dig out more controllable ability for text-to-image diffusion models.
\newblock \emph{arXiv preprint arXiv:2302.08453}, 2023.

\bibitem[Peebles and Xie(2023)]{peebles2023scalable}
William Peebles and Saining Xie.
\newblock Scalable diffusion models with transformers.
\newblock In \emph{Proceedings of the IEEE/CVF International Conference on Computer Vision}, pages 4195--4205, 2023.

\bibitem[Qi et~al.(2017)Qi, Su, Mo, and Guibas]{qi2017pointnet}
Charles~R Qi, Hao Su, Kaichun Mo, and Leonidas~J Guibas.
\newblock Pointnet: Deep learning on point sets for 3d classification and segmentation.
\newblock In \emph{Proceedings of the IEEE conference on computer vision and pattern recognition}, pages 652--660, 2017.

\bibitem[Radford et~al.(2019)Radford, Wu, Child, Luan, Amodei, Sutskever, et~al.]{radford2019language}
Alec Radford, Jeffrey Wu, Rewon Child, David Luan, Dario Amodei, Ilya Sutskever, et~al.
\newblock Language models are unsupervised multitask learners.
\newblock \emph{OpenAI blog}, 1\penalty0 (8):\penalty0 9, 2019.

\bibitem[Rombach et~al.(2021)Rombach, Blattmann, Lorenz, Esser, and Ommer]{rombach2021highresolution}
Robin Rombach, Andreas Blattmann, Dominik Lorenz, Patrick Esser, and Björn Ommer.
\newblock High-resolution image synthesis with latent diffusion models, 2021.

\bibitem[Saharia et~al.(2022)Saharia, Chan, Saxena, Li, Whang, Denton, Ghasemipour, Gontijo~Lopes, Karagol~Ayan, Salimans, et~al.]{saharia2022photorealistic}
Chitwan Saharia, William Chan, Saurabh Saxena, Lala Li, Jay Whang, Emily~L Denton, Kamyar Ghasemipour, Raphael Gontijo~Lopes, Burcu Karagol~Ayan, Tim Salimans, et~al.
\newblock Photorealistic text-to-image diffusion models with deep language understanding.
\newblock \emph{Advances in neural information processing systems}, 35:\penalty0 36479--36494, 2022.

\bibitem[Sallab et~al.(2019)Sallab, Sobh, Zahran, and Essam]{sallab2019lidar}
Ahmad~El Sallab, Ibrahim Sobh, Mohamed Zahran, and Nader Essam.
\newblock Lidar sensor modeling and data augmentation with gans for autonomous driving.
\newblock \emph{arXiv preprint arXiv:1905.07290}, 2019.

\bibitem[Song et~al.(2021)Song, Sohl-Dickstein, Kingma, Kumar, Ermon, and Poole]{song2020score}
Yang Song, Jascha Sohl-Dickstein, Diederik~P Kingma, Abhishek Kumar, Stefano Ermon, and Ben Poole.
\newblock Score-based generative modeling through stochastic differential equations.
\newblock In \emph{International Conference on Learning Representations}, 2021.

\bibitem[Swerdlow et~al.(2024)Swerdlow, Xu, and Zhou]{swerdlow2024street}
Alexander Swerdlow, Runsheng Xu, and Bolei Zhou.
\newblock Street-view image generation from a bird's-eye view layout.
\newblock \emph{IEEE Robotics and Automation Letters}, 2024.

\bibitem[Touvron et~al.(2023)Touvron, Lavril, Izacard, Martinet, Lachaux, Lacroix, Rozi{\`e}re, Goyal, Hambro, Azhar, et~al.]{touvron2023llama}
Hugo Touvron, Thibaut Lavril, Gautier Izacard, Xavier Martinet, Marie-Anne Lachaux, Timoth{\'e}e Lacroix, Baptiste Rozi{\`e}re, Naman Goyal, Eric Hambro, Faisal Azhar, et~al.
\newblock Llama: Open and efficient foundation language models.
\newblock \emph{arXiv preprint arXiv:2302.13971}, 2023.

\bibitem[Unterthiner et~al.(2018)Unterthiner, van Steenkiste, Kurach, Marinier, Michalski, and Gelly]{unterthiner2018towards}
Thomas Unterthiner, Sjoerd van Steenkiste, Karol Kurach, Raphael Marinier, Marcin Michalski, and Sylvain Gelly.
\newblock Towards accurate generative models of video: A new metric \& challenges.
\newblock \emph{arXiv preprint arXiv:1812.01717}, 2018.

\bibitem[Van Den~Oord et~al.(2017)Van Den~Oord, Vinyals, et~al.]{van2017vqvae}
Aaron Van Den~Oord, Oriol Vinyals, et~al.
\newblock Neural discrete representation learning.
\newblock \emph{Advances in neural information processing systems}, 30, 2017.

\bibitem[Vaswani et~al.(2017)Vaswani, Shazeer, Parmar, Uszkoreit, Jones, Gomez, Kaiser, and Polosukhin]{vaswani2017attention}
Ashish Vaswani, Noam Shazeer, Niki Parmar, Jakob Uszkoreit, Llion Jones, Aidan~N Gomez, {\L}ukasz Kaiser, and Illia Polosukhin.
\newblock Attention is all you need.
\newblock \emph{Advances in neural information processing systems}, 30, 2017.

\bibitem[Wang et~al.(2023)Wang, Zhu, Huang, Chen, and Lu]{wang2023drivedreamer}
Xiaofeng Wang, Zheng Zhu, Guan Huang, Xinze Chen, and Jiwen Lu.
\newblock Drivedreamer: Towards real-world-driven world models for autonomous driving.
\newblock \emph{arXiv preprint arXiv:2309.09777}, 2023.

\bibitem[Wang et~al.(2024)Wang, He, Fan, Li, Chen, and Zhang]{wang2024driving}
Yuqi Wang, Jiawei He, Lue Fan, Hongxin Li, Yuntao Chen, and Zhaoxiang Zhang.
\newblock Driving into the future: Multiview visual forecasting and planning with world model for autonomous driving.
\newblock In \emph{2024 IEEE/CVF Conference on Computer Vision and Pattern Recognition (CVPR)}. IEEE, 2024.

\bibitem[Weng et~al.(2021)Weng, Wang, Levine, Kitani, and Rhinehart]{weng2021SPFNet}
Xinshuo Weng, Jianren Wang, Sergey Levine, Kris Kitani, and Nicholas Rhinehart.
\newblock Inverting the pose forecasting pipeline with spf2: Sequential pointcloud forecasting for sequential pose forecasting.
\newblock In \emph{Conference on robot learning}, pages 11--20. PMLR, 2021.

\bibitem[Weng et~al.(2022)Weng, Nan, Lee, McAllister, Gaidon, Rhinehart, and Kitani]{weng2022s2net}
Xinshuo Weng, Junyu Nan, Kuan-Hui Lee, Rowan McAllister, Adrien Gaidon, Nicholas Rhinehart, and Kris~M Kitani.
\newblock S2net: Stochastic sequential pointcloud forecasting.
\newblock In \emph{European Conference on Computer Vision}, pages 549--564. Springer, 2022.

\bibitem[Wilson et~al.(2021)Wilson, Qi, Agarwal, Lambert, Singh, Khandelwal, Pan, Kumar, Hartnett, Kaesemodel~Pontes, Ramanan, Carr, and Hays]{NEURIPSDATASETSANDBENCHMARKS2021_4734ba6f}
Benjamin Wilson, William Qi, Tanmay Agarwal, John Lambert, Jagjeet Singh, Siddhesh Khandelwal, Bowen Pan, Ratnesh Kumar, Andrew Hartnett, Jhony Kaesemodel~Pontes, Deva Ramanan, Peter Carr, and James Hays.
\newblock Argoverse 2: Next generation datasets for self-driving perception and forecasting.
\newblock In \emph{Proceedings of the Neural Information Processing Systems Track on Datasets and Benchmarks}, 2021.

\bibitem[Xiong et~al.(2023)Xiong, Ma, Wang, and Urtasun]{xiong2023ultralidar}
Yuwen Xiong, Wei-Chiu Ma, Jingkang Wang, and Raquel Urtasun.
\newblock Learning compact representations for lidar completion and generation.
\newblock In \emph{Proceedings of the IEEE/CVF Conference on Computer Vision and Pattern Recognition}, pages 1074--1083, 2023.

\bibitem[Yang et~al.(2024)Yang, Gao, Qiu, Chen, Li, Dai, Chitta, Wu, Zeng, Luo, Zhang, Geiger, Qiao, and Li]{yang2024genad}
Jiazhi Yang, Shenyuan Gao, Yihang Qiu, Li Chen, Tianyu Li, Bo Dai, Kashyap Chitta, Penghao Wu, Jia Zeng, Ping Luo, Jun Zhang, Andreas Geiger, Yu Qiao, and Hongyang Li.
\newblock {Generalized Predictive Model for Autonomous Driving}.
\newblock In \emph{Proceedings of the IEEE/CVF Conference on Computer Vision and Pattern Recognition (CVPR)}, 2024.

\bibitem[Yang et~al.(2023)Yang, Ma, Peng, Guo, Lin, and Yu]{yang2023bevcontrol}
Kairui Yang, Enhui Ma, Jibin Peng, Qing Guo, Di Lin, and Kaicheng Yu.
\newblock Bevcontrol: Accurately controlling street-view elements with multi-perspective consistency via bev sketch layout.
\newblock \emph{arXiv preprint arXiv:2308.01661}, 2023.

\bibitem[Yumeng et~al.(2024)Yumeng, Shi, Kaixin, Xiaoqing, Xiao, Fan, Jizhou, Hua, and Haifeng]{zhang2024bevworld}
Zhang Yumeng, Gong Shi, Xiong Kaixin, Ye Xiaoqing, Tan Xiao, Wang Fan, Huang Jizhou, Wu Hua, and Wang Haifeng.
\newblock Bevworld: A multimodal world model for autonomous driving via unified bev latent space.
\newblock \emph{arXiv preprint arXiv:2407.05679}, 2024.

\bibitem[Zhang et~al.(2023)Zhang, Xiong, Yang, Casas, Hu, and Urtasun]{zhang2023copilot4d}
Lunjun Zhang, Yuwen Xiong, Ze Yang, Sergio Casas, Rui Hu, and Raquel Urtasun.
\newblock Learning unsupervised world models for autonomous driving via discrete diffusion.
\newblock \emph{arXiv preprint arXiv:2311.01017}, 2023.

\bibitem[Zhao et~al.(2024)Zhao, Wang, Zhu, Chen, Huang, Bao, and Wang]{zhao2024drivedreamer}
Guosheng Zhao, Xiaofeng Wang, Zheng Zhu, Xinze Chen, Guan Huang, Xiaoyi Bao, and Xingang Wang.
\newblock Drivedreamer-2: Llm-enhanced world models for diverse driving video generation.
\newblock \emph{arXiv preprint arXiv:2403.06845}, 2024.

\bibitem[Zyrianov et~al.(2022)Zyrianov, Zhu, and Wang]{zyrianov2022lidargen}
Vlas Zyrianov, Xiyue Zhu, and Shenlong Wang.
\newblock Learning to generate realistic lidar point clouds.
\newblock In \emph{European Conference on Computer Vision}, pages 17--35. Springer, 2022.

\end{thebibliography}
